\documentclass[conference]{IEEEtran}
\IEEEoverridecommandlockouts

\AtBeginDocument{%
  \providecommand\BibTeX{{%
    \normalfont B\kern-0.5em{\scshape i\kern-0.25em b}\kern-0.8em\TeX}}}

\usepackage{hyperref}
\usepackage{color,soul,xspace}
\usepackage{amsmath,amssymb}
\usepackage[noend]{algpseudocode}
\usepackage{multirow}
\usepackage{tabu}
\usepackage{caption}
\usepackage{algorithm}
\usepackage{float}
\usepackage{graphicx}
\usepackage{subcaption}
\usepackage{emp}
\newcommand{\etal}{\textit{et al.}}
\newcommand{\mourmeth}{\text{UISS}}
\newcommand{\ourmeth}{$\mourmeth$\xspace}

\newcommand{\vsa}{\vspace*{-0cm}}
\newcommand{\vsb}{\vspace*{-0cm}}

\def\BibTeX{{\rm B\kern-.05em{\sc i\kern-.025em b}\kern-.08em
    T\kern-.1667em\lower.7ex\hbox{E}\kern-.125emX}}

\title{Unsupervised Instance and Subnetwork Selection for Network Data}

\author{\IEEEauthorblockN{Lin Zhang}
\IEEEauthorblockA{\textit{International Digital} \\ \textit{Economy Academy (IDEA)} \\
Shenzhen, China \\
zhanglin@idea.edu.cn}
\and
\IEEEauthorblockN{Nicholas Moskwa}
\IEEEauthorblockA{\textit{Biological Sciences}\\ 
\textit{University at Albany - SUNY} \\
New York, USA \\
nmoskwa@albany.edu}
\and
\IEEEauthorblockN{Melinda  Larsen}
\IEEEauthorblockA{\textit{Biological Sciences}\\ \textit{University at Albany - SUNY} \\
New York, USA \\
mlarsen@albany.edu}
\and
\IEEEauthorblockN{Petko Bogdanov}
\IEEEauthorblockA{\textit{Computer Science}\\ \textit{University at Albany - SUNY} \\
New York, USA \\
pbogdanov@albany.edu}
}

\begin{document}

\maketitle
\begin{abstract}
Unlike tabular data, features in network data are interconnected within a domain-specific graph. Examples of this setting include gene expression overlaid on a protein interaction network (PPI) 
and user opinions in a social network. Network data is typically high-dimensional (large number of nodes) and often contains outlier snapshot instances and noise. In addition, it is often non-trivial and time-consuming to annotate instances with global labels (e.g., disease or normal). How can we jointly select discriminative subnetworks and representative instances for network data without supervision?  

We address these challenges within an unsupervised framework for joint subnetwork and instance selection in network data, called \ourmeth, via a convex self-representation objective.
Given an unlabeled network dataset, \ourmeth identifies representative instances while ignoring outliers. It outperforms state-of-the-art baselines on both discriminative subnetwork selection and representative instance selection, achieving up to $10\%$ accuracy improvement on all real-world data sets we use for evaluation. 
When employed for exploratory analysis in RNA-seq network samples from multiple studies it produces interpretable and informative summaries.
\end{abstract}


\section{Introduction}

Network data abounds 
in a wide range of application domains: from activation snapshots of the human brain to the global state of sensor and online social networks.
Different from tabular data, network data includes a shared structure associating features (nodes) in addition to their values. This structured feature space enables robust and interpretable solutions in a host of tasks, such as subnetwork selection~\cite{DIPS,ranu2013mining}, graph signal decomposition~\cite{mcneil2021tgsd}, network-aware distance measures~\cite{amelkin2017distance} and summarization~\cite{SilvaICDE2015,mcneil2022saga}.

\begin{figure}[t]
    \centering
    \includegraphics[width=0.45\textwidth]{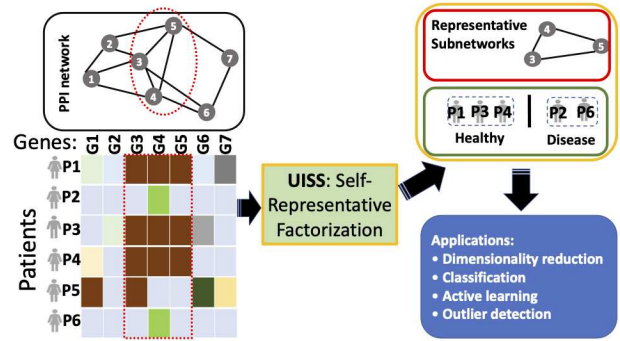}
    \caption{ 
    An illustrative example for unsupervised instance and subnetwork learning from a set of instances---patients' gene expression, sharing a common network structure---the human protein-protein interaction network (PPI) associating genes of known interactions. 
    Our goal is to jointly select representative instances and discriminative subnetworks which can then be used for various downstream tasks.
    } \vsa\vsb
    \label{fig:overview}
\end{figure}

A network sample (i.e., instance) consists of feature values (node weights) and a structure shared by all instances. Consider the example in Fig.~\ref{fig:overview} of gene expression for $6$ patients ($P1-P6$), where the expression levels for each gene ($G1-G7$) are employed as features and the human interactome is the structure associating the features. 
Our goal is to select a subspace of well-connected subnetworks which also ``distinguish'' the natural instance clusters. One such feature set in the example includes genes $\{G3, G4, G5\}$ forming a connected PPI subnetwork which also elucidates well-pronounced clusters of patients: $\{P1,P3,P4\}$ and $\{P2,P6\}$. Another goal is to identify outliers, such as patient $P5$ who does not ``align'' well with the natural clusters.

It is important to note that network instance data is different from network structure data employed for structural subgraph mining and classification since the latter employs subgraph occurrences as features as opposed to node values within a fixed structure~\cite{jiang2013fsgsurvey}. 
Our setting is also different from collective classification~\cite{sen2008collective} typically arising in social media mining~\cite{Cheng:2017:UFS:3097983.3098106}, 
where individual nodes are instances as opposed to the whole network snapshot. 

One important task in network instance data analysis is the prediction of the global state (label) of network instances~\cite{DIPS,Zhang2019dsl} with applications in neuroscience~\cite{bogdanov2017learning},
gene expression on protein-protein interaction (PPI) networks~\cite{dutkowski2011protein} and analysis of the state of transportation and sensor networks~\cite{bms11}. Existing work typically focuses on the supervised setting, where each sample is characterized by a global label (e.g., disease/healthy). To tackle the typical high dimensionality 
existing work exploits the network structure to identify connected discriminative sub-spaces (subgraphs and associated feature values) to construct classifiers which generalize and offer interpretability~\cite{DIPS,ranu2013mining}.

While it may be easy to obtain a network instance, obtaining reliable global labels for supervision may be challenging 
due to expensive human-in-the-loop tests (e.g., doctor assessment for cognitive impairments~\cite{ADNI}), or complex global states (e.g., regimes of operation of infrastructure networks such as road networks and power grids). In such scenarios unsupervised methods can both elucidate important subnetworks and representative instances to be labeled (akin to active learning~\cite{Settles10activelearning}), while avoiding outlier and redundant samples. 
We propose \ourmeth, an unsupervised approach to jointly learn discriminative subgraphs  and representative instances for network data. We formalize the problem as a self-representative factorization of the data which also maximizes the connectivity of selected nodes and the discriminative power of selected instances. In particular, we select subnetworks and instances that can reconstruct the full data (self-representation) 
and discount loss due to outlier instances. We enforce network locality for selected nodes by a quadratic penalty based on the network Laplacian. Finally, the representative power of selected instances is promoted via an orthogonality constraint. 

The main contributions of this paper are as follows:\\
\noindent{\bf 1. Significance:} We address an unexplored problem of unsupervised learning for network samples by jointly detecting discriminative subgraphs and representative instances. 

\noindent{\bf 2. Robustness:} \ourmeth is robust to noise and outliers, outperforming baselines in feature and instance selection and achieving up to $10\%$ accuracy improvement in real-world data. 

\noindent{\bf 3. Interpretability and applicability:} \ourmeth discovers interpretable subspaces such as discriminative city-center localities in bike rental behavior, and reconstructs the developmental timeline in RNA-seq samples from organoid development. 

\section{Related Work}

\noindent{\bf Unsupervised feature selection:} Methods in this category employ filtering~\cite{He:2005:LSF:2976248.2976312} 
and embedding~\cite{Yang:2011:LNR:2283516.2283660},
however, the resulting features are not guaranteed to be discriminative for the actual (withheld) instance classes. 
Learning discriminative features based on cluster (pseudo) labels have been proposed when actual instance labels are absent~\cite{Li:2012:UFS:2900728.2900874}. 
Other discriminative feature selection methods~\cite{Qian:2013:RUF:2540128.2540361,Li2016RobustUF,Du:2015:UFS:2783258.2783345} 
directly exploit cluster information: UDFS~\cite{Yang:2011:LNR:2283516.2283660} maximizes the cluster separation and compactness; RUFS~\cite{Qian:2013:RUF:2540128.2540361} learns jointly clusters and feature subsets by non-negative matrix factorization; NFDS~\cite{Li:2012:UFS:2900728.2900874} imposes orthogonality constraints to avoid cluster ambiguity; and MCFS~\cite{Cai:2010:UFS:1835804.1835848} preserves across-cluster structure within a spectral embedding.
Gu~\etal~\cite{Gu2011IJCAI} developed feature selection method using the structure among instances.
While all above methods optimize the discriminative power of features, (i) they cannot select subnetworks in network samples,
(ii) they are not robust to outliers which have been shown to impact unsupervised learning~\cite{Liu:2010:RGM:3104322.3104408}; 
and (iii) they expect the number of clusters as an input parameter. 
We overcome these shortcomings by combining network smoothness and cluster-representative instance selection via an orthogonality constraint and by self-representation.

Feature selection in networked settings has also been proposed for cases when the instances are associated within a known network~\cite{li2019kdd,wei2016aaai,li2015cikm,wei2015sdm}. However, this setting is different (and complementary) to ours in that the network structure associates instances as opposed to features.

\noindent{\bf Instance and feature co-selection:} Feature and instance selection are co-dependent tasks since representative instances highlight relevant features and vice versa. 
CoSelect~\cite{coselection} and gActive~\cite{Kong:2011:DAF:2020408.2020511} learn features and instances effectively but require global label information and are applicable in supervised settings only. 
Li \etal~\cite{Li2018alfs} developed an alternative unsupervised model, called ALFS which, however, does not consider a network structure among features or the existence of outliers, rendering it less robust than our solution as we demonstrate empirically. 
Our settings also differ from supervised
active learning~\cite{Settles10activelearning}, which has been widely employed in text~\cite{Tong:2002:SVM:944790.944793}
and network data~\cite{coselection}. 
Active learning in a fully unsupervised setting has been pursued via 
transductive experimental design (TED)~\cite{Yu:2006:ALV:1143844.1143980}
which performs instance selection guided by reconstruction error. 
Following a similar minimum-reconstruction-error approach, ALFS~\cite{Li2018alfs} was proposed to jointly select representative instances and features. 

\noindent{\bf Learning with network data:} Closest to our setting are recent supervised methods for labeled network data~\cite{DIPS,DSR,ranu2013mining,Zhang2019dsl}. 
They adopt sampling or optimization to  
learn subgraphs with high discriminative power for supervised network state prediction. 
Unsupervised learning for network data, however, presents a new challenge which has not been studied in the literature to the best of our knowledge. 


\section{Preliminaries}

A network instance ${\mathcal{S}}_i = \left \{ V_i , E_i, { {\bf X}_i} \right \}$, is the triplet containing nodes $V_i$, edges $E_i$ and node/feature values $ {\bf X}_i$. A dataset ${\mathcal{DS}}=\{S_1\dots S_n\}$ is a set of $n$ network instances. All instances share the same network structure.
Edges in ${\mathcal{S}}_i$ are weighted by the fraction of their occurrences in instances ${\bf M}_{pq} =1/n\sum_i E_i(p,q)$. 
The graph Laplacian matrix of ${\mathcal{S}}_i$ is defined as ${\bf L} =  { \bf D}-{\bf M}$, where
$ {\bf D}$ is the diagonal matrix of weighted node degrees, i.e., ${\bf D}_{pp} = \sum_{q}{\bf M}_{pq}$. The data matrix $\mathbf{X} = \left [ {X}_1,{X}_2,...{X}_n \right ]^T \in \mathbb{R}^{n\times m}$ of a dataset ${\mathcal{DS}}$ is comprised of network instance values ${X}_i$ in its rows.

\section{Problem Formulation}

Given a dataset ${\mathcal{DS}}$ of $n$ unlabeled network instances with $m$ nodes, one of our goal is to select $p~(p\ll n)$ instances which are representative of the underlying clusters in the dataset and also to discover a connected subgraph subspace of $q~(q\ll m)$ nodes. Intuitively, the above design objective assumes that some instances form (an unknown number of) clusters, while others are outliers and do not belong to clusters. 



\noindent{\bf Self-representative factorization:} We model the joint instance and subgraph selection as a sparse self-representative factorization of the observations $\mathbf{X}$. 
In traditional matrix factorization $\mathbf{X}=\mathbf{U}\mathbf{V}$~\cite{Ding06kdd}, the two factors can be thought of as latent indicators of row and column clusters. Such representations are efficient and compact, but are also shown to exchange points from separate low-dimensional structures in the data, leading to sub-optimal clustering performance~\cite{Mahoney2009}. Instead, we propose a self-representative reconstruction based on the principle that a matrix can be represented by factors selected from its own rows and columns.
In particular, we introduce selectors for features ${\bf P}\in \mathbb{R}^{m\times k}$ and instances ${\bf Q}\in \mathbb{R}^{k\times n}$, leading to:
\begin{equation}\label{objective}
        \begin{aligned}
    &\underset{{\bf P,Q}}{\mathrm{argmin}} \left \|\mathbf{X}-\mathbf{UV} \right \|_{2,1}
        + \lambda_1\left \| \mathbf{P} \right \|_{2,1}+\lambda_2\left \| \mathbf{Q}^T  \right \|_{2,1},\\
        & ~~~~s.t.~\mathbf{U} = \mathbf{XP}, \mathbf{V} = \mathbf{QX},  \mathbf{P}\geq 0, \mathbf{Q}\geq 0,
   \end{aligned}  
\end{equation}
where the $L_{2,1}$ norm is employed to ensure robustness to outliers and noise through balance parameters $\lambda_1$ and $\lambda_2$. 
The key distinction from existing latent factorization is that factors are selected from the data columns and rows themselves. One can also view our co-selection strategy as imposing a reduced-rank structure on a CUR-like decomposition~\cite{Mahoney2009} via a bi-linear factorization employing $\bf P$ and $\bf Q$. 
Such a low-rank structure promotes a compact and interpretable latent space of the instances and subnetworks. Note that subnetwork and instance selection is based on thresholding $\bf P$'s rows and  $\bf Q$'s columns rather than a fixed number of components pre-specified by the user. In particular, the importance of subnetworks and instances can be quantified based on the corresponding row norm of $\bf P$ and column norm of $\bf Q$.

\noindent{\bf Discriminative power using an orthogonal instance subspace:} So far our objective ensures instance and subnetwork selection, but does not ensure that the selected instances are representative of the underlying clusters as opposed to outliers. 
Instead of explicitly modeling cluster memberships and deciding on the number of clusters apriori, we enforce orthogonality on the selected instances via a constraint of the form $\mathbf{V}\mathbf{V}^T = \mathbf{I}$. 
Columns of $\bf V$ act as cluster indicators, but also as a dictionary basis to reconstruct all instances, and thus
this orthogonal constraint within the factorization ensures separation of selected instances. While the important role of orthogonality in traditional factorization clustering has long been recognized~\cite{Ding06kdd}, we impose orthogonality within a self-representative factorization, leading to a superior performance in both feature and instance selection for network data.    

\noindent{\bf Graph connectivity:} We also impose ``smoothness'' of the subnetwork selection with respect to the common graph structure $\mathcal{S}$ to encourage connectivity in the feature space via a trace norm: $\textup{Tr}(\mathbf{P}^T \mathbf{L}\mathbf{P})$, where $\mathbf{L}$ is the graph Laplacian matrix of the nodes' network.

Combining all modeling goals 
we obtain:
\begin{equation}\label{objective} 
        \begin{aligned}
    &\underset{{\bf P,Q}}{\mathrm{argmin}} \left \|\mathbf{X}-\mathbf{UV} \right \|_{2,1}
        + \lambda_1\left \| \mathbf{P} \right \|_{2,1}+\lambda_2\left \| \mathbf{Q}^T  \right \|_{2,1} +
        \lambda_3 \textup{Tr}(\mathbf{P}^T \mathbf{L}\mathbf{P}),\\
        & ~s.t.~\mathbf{U} = \mathbf{XP}, \mathbf{V} = \mathbf{QX}, \mathbf{V}\mathbf{V}^T = \mathbf{I},
        \mathbf{P},\mathbf{Q}\geq 0.\vsb
   \end{aligned}  
\end{equation}
The objective unifies our design goals: joint (i) subnetwork selection via $\mathbf{P}$, and (ii) discriminative instance selection via $\mathbf{Q}$, where discriminative power is promoted by $\mathbf{V}$'s orthogonality. 




\section{Optimization Algorithm}
Our objective from Eq.~\ref{objective} is jointly convex, however, it is non-trivial to develop gradient-based solutions for it directly 
because the $L_{2,1}$ norm is non-smooth.
Instead, we propose an alternating optimization algorithm which efficiently updates the instance selector $\bf P$ and feature selector $\bf Q$ in turn until convergence.


{\noindent \bf Update of Q:}
To update $\bf Q$, we fix $\bf P$ and obtain:
\begin{equation}\label{update_QER}
        \begin{aligned}
    &\underset{{\bf V,Q}}{\mathrm{argmin}} \hspace{0.05cm}\left \|{{\bf X}}-\mathbf{U}{\bf V}\right \|_{2,1}
        +\left \| {\bf V}-{\bf QX} \right \|_F^2 +\lambda_2\left \| \mathbf{Q}^T  \right \|_{2,1},\\
         &~~~ s.t.~\mathbf{V}\mathbf{V}^T = \mathbf{I}, \mathbf{Q}\geq 0 \vsb
   \end{aligned}  
\end{equation}
We set the gradient w.r.t $\bf Q$ of Eq.~\ref{update_QER} to zero:
\begin{equation}\label{update_R}
        \begin{aligned}
({\bf QX}-{\bf V}){\bf X}^T +\lambda_2 {\bf Q}{\bf \Pi }=\mathbf{0}, 
   \end{aligned}  
\end{equation}
where $\bf \Pi$ is a diagonal matrix: 
${\bf \Pi}_{ii} =(2\left \| \mathbf{Q}^i \right \|_2)^{-1}$. Thus, we obtain a closed-form solution for Eq.~\ref{update_R}:  
${\bf{Q}} = {\bf V}{\bf X}^T({\bf X}{\bf X}^T+\lambda_2{\bf \Pi})^{-1}$. Note that $({\bf X}{\bf X}^T+\lambda_2{\bf \Pi})$ is invertible since ${\bf X}^T{\bf X}$ is positive definite and ${\bf \Pi}$ is a non-negative diagonal matrix. 

To handle the orthogonality constraint in Eq.~\ref{update_QER},
we introduce an intermediate variable $\bf{W}$ which approximates $\bf{V}$, leading to two subproblems:
\begin{equation}~\label{RQ_solver_02}
\begin{aligned}
\begin{cases}
\underset{{ \mathbf{V}}}{\mathrm{argmin}} \hspace{0.05cm}\left \|{{\bf X}}-\mathbf{U}{\bf V}\right \|_{2,1}
        +\left \| {\bf V}-{\bf QX} \right \|_F^2 +\left \| {\bf V}-{\bf W} \right \|_F^2\textup{(a)}\\
\underset{{\bf W}}{\mathrm{argmin}} \hspace{0.05cm}\left \| {\bf V}-{\bf W} \right \|_F^2,\hspace{0.05cm} s.t \hspace{0.1cm}\mathbf{W}\mathbf{W}^T = \mathbf{I}~~~~~~~~~~~~~~~~~~~~~~\textup{(b)}
\end{cases}
   \end{aligned} 
\end{equation}
We set the gradient w.r.t. $\bf V$ in Eq.~\ref{RQ_solver_02}(a) to $\mathbf{0}$:
\begin{equation}\label{SOLVER_V}
        \begin{aligned}
   {\bf U}^T({\bf U}{\bf V}-{{\bf X}}){\bf \Theta } +{\bf V}-{\bf QX} + {\bf V}-{\bf W}= \mathbf{0},
   \end{aligned}  
\end{equation}
where  ${\bf \Theta}$ is diagonal with elements ${\bf \Theta}_{ii} = \frac{1}{ 2\left \| {{\bf X}}^i-{\bf U}{\bf V}^i  \right \|_2}$ when ${{\bf X}}^i-{\bf U}{\bf V}^i \neq \mathbf{0}$, and $\mathbf{0}$ otherwise. 
Since ${\bf U}^T{\bf U}$ is symmetric and positive semi-definite, its eigendecomposition
${\bf U}^T{\bf U} = {\bf A} {\bf \Sigma } {\bf A}^T$ has real eigenvalues (diagonal of ${\bf \Sigma }$) and real orthonormal eigenvectors in {\bf A}.  
Employing this decomposition, we can rewrite (\ref{SOLVER_V}) as 
\begin{equation}~\label{Q_solover022}
     \begin{aligned}
    {\bf A} {\bf \Sigma } {\bf A}^T{\bf V}  {\bf \Theta} +2{\bf V}= {{{\bf \Psi}}},
     \end{aligned}  
\end{equation}
where ${{\bf \Psi}} = {\bf U}^T{{\bf X}}{{\bf \Theta}}+{\bf QX} + {\bf W} $. 
Since $\bf{ A}$ is orthonormal, we can multiply Eq.~\ref{Q_solover022} by ${\bf A}^T$ on the left side, which after substituting ${{{\bf E } } } = {\bf A}^T{\bf V}$ can be simplified to $     {\bf \Sigma } {{{\bf E } } } {\bf \Theta } +2{{{\bf E}}}= {\bf A}^T{{{\bf \Psi}}}$. 
Individual elements of $ {{\bf E } }$ can then be updated as follows
: $    {{{\bf E } } }_{ij} = {\left [ {\bf A}^T{{{\bf \Psi}}} \right ]}_{ij}/ ({\bf \Sigma }_{ii} {\bf \Theta }_{jj}+ 2)$
and $\bf V$ can be obtained as ${\bf V} = {\bf A} {{{\bf E } }}$.

There exists a closed-form solution $\mathbf{W} = \mathbf{R} \mathbf{I} \mathbf{T}^T$ solution for Eq.~\ref{RQ_solver_02}(b) due to Viklands \etal~\cite{Viklands2006AlgorithmsFT}, where $\mathbf{R}$ and $\mathbf{T}$ are the left and right singular vectors of $\bf V$ in an SVD decomposition. 


{\noindent \bf{Update of {\bf P}}: }The subproblem w.r.t $\bf P$ 
becomes: 
\begin{equation*}~\label{solve_P}
        \begin{aligned}
    &\underset{{\bf P}}{\mathrm{argmin}}\hspace{0.05cm} \left \|{{\bf X}}-\mathbf{X}{\bf P}\mathbf{V}\right \|_{2,1}
        + \lambda_1\left \| \mathbf{P} \right \|_{2,1}
        + \lambda_3 \textup{Tr}(\mathbf{P}^T \mathbf{L}\mathbf{P})
   \end{aligned}  
\end{equation*}
We again employ an auxiliary
${\bf U}$ to approximate $\mathbf{X}{\bf P}$:
\begin{equation}~\label{PT_solver_00}
        \begin{aligned}
        &\underset{{\bf P,U}}{\mathrm{argmin}}\hspace{0.05cm} 
         \left \|{{\bf X}}-\mathbf{U}\mathbf{X}\right \|_{2,1}
        + \lambda_1\left \| \mathbf{P} \right \|_{2,1}
        + \lambda_3 \textup{Tr}(\mathbf{P}^T \mathbf{L}\mathbf{P}) \\ &
        + \left \| \mathbf{U}-\mathbf{XP} \right \|_F^2,
   \end{aligned}  
\end{equation}
and alternate between $\bf P$ and $\bf U$ updates.
We set the gradient w.r.t. $\bf U$ in Eq.~\ref{PT_solver_00} to $\bf 0$:
\begin{equation}
    \begin{aligned}
\mathbf{K}(\mathbf{U}\mathbf{V}-{{\bf X}})\mathbf{V}^T + \mathbf{U}-\mathbf{XP}=0,
    \end{aligned}
    \label{eq:gradU}
\end{equation}
where $\bf{K}$ is a diagonal matrix with elements
${\bf K}_{ii} = \left ( 2\left \| {{\bf X}}_i-{\bf U}_i {\bf V}  \right \|_2 \right )^{-1}$ when ${{\bf X}}_i-{\bf U}_i {\bf X}\neq 0$, and ${\bf K}_{ii}=0$ otherwise. 
Similar to the solution for $\bf{V}$, we employ the eigendecomposition of ${\bf V}{\bf V}^T = {\bf B} {\bf \Lambda } {\bf B}^T$ and multiply by ${\bf B}$ on the right, thus simplifying Eq.~\ref{eq:gradU} to $  \mathbf{K}\mathbf{H} {\bf \Lambda }  + {\bf H }=  {\bf \Xi }{\bf B}$, 
where  ${\bf H} = {\bf UB}$ and  ${\bf \Xi } = \mathbf{XP}+ \mathbf{K}{{\bf X}}{\bf V}^T$. 
After updating the elements of ${\bf H}$ according to ${\bf H}_{ij} = \frac{[{\bf \Xi B}]_{ij}}{{\bf K}_{ii}{\bf \Lambda}_{jj}+1 }$, we get a closed-form solution for $\bf{U}$: ${\bf U} = {\bf HB}^T$. 
We set the gradient w.r.t. $\bf{P}$ in Eq.~\ref{PT_solver_00} to $\bf 0$: 
\begin{equation}~\label{solve_P14}
        \begin{aligned}
        {\bf{X}}^T\left ( \mathbf{XP}-\mathbf{U} \right ) 
        +\lambda_1{\bf{G}} \mathbf{P}
        + \lambda_3 \mathbf{L}\mathbf{P}=0,
   \end{aligned}  
\end{equation}
where $\bf{G}$ is diagonal with elements ${\bf G}_{ii}=(2\left \| \mathbf{P }_i \right \|_2)^{-1}$ when  $\left \|\mathbf{P }_i\right \|_2\neq 0$, and ${\bf G}_{ii}= 0$ otherwise.
Thus, $\bf P$ has the following closed-form solution:
\begin{equation}~\label{update_P}
    \begin{aligned}
 {\bf P} =  (\mathbf{X}^T\mathbf{X} + \lambda_1\mathbf{G}+\lambda_3 \mathbf{L})^{-1} {\bf{X}}^T \mathbf{U}, 
    \end{aligned}
\end{equation}
where $\mathbf{X}^T\mathbf{X} + \lambda_1\mathbf{G}+\lambda_3 \mathbf{L}$ is invertible due to all summands being positive semi-definite matrices.





\begin{center}
	\begin{algorithm}[!tp]
		\caption{\ourmeth} 
		\label{alg:opt}
		\begin{algorithmic}[1]
			\Require{Training data $\bf{X}$, and parameters({$\lambda_1,\lambda_2,\lambda_3, k$) 
			}}
			\Ensure {Selection matrices $\bf P,Q$}
			\State Initialize $\bf P,Q$, ${\bf \Theta}$, ${\bf \Pi}$ and ${\bf K}$ to identity matrices; 
			 $\bf{U}$, $\bf V$, to random matrices;
		    \While{$\bf P,Q$ not converged}
		        \State  $\left ( {\bf A} ,{\bf \Sigma }   \right ) = evd\left ( {\bf U}^T{\bf U} \right )$
		        \State ${\bf \Theta}_{ii} = \left ( 2\left \| {{\bf X}}^i-{\bf U}{\bf V}^i  \right \|_2 \right )^{-1}$  if ${{\bf X}}^i-{\bf U}{\bf V}^i \neq \mathbf{0}$
		        \While{$\bf{W}$ and $\bf{V}$ have not converged}
		            \State ${{\bf \Psi}} = {\bf U}^T{{\bf X}}{{\bf \Theta}}+{\bf QX} + {\bf W} $
                    \State ${{{\bf E } } }_{ij} = {\left [ {\bf A}^T{{{\bf \Psi}}} \right ]}_{ij}/ ({\bf \Sigma }_{ii} {\bf \Theta }_{jj}+ 2)$ 
                    \State ${\bf V} = {\bf A} {\bf E }$
                    \State $({\mathbf{R}},{\mathbf{T}})= svd \left(\mathbf{V}\right)$
                    \State ${\bf W} ={\bf R}{\bf I}{\bf T}^T $
                \EndWhile
                \State $ {\bf{Q}} = max[{\bf V}{\bf X}^T({\bf X}{\bf X}^T+ \lambda_2{\bf \Pi})^{-1},0]$
                \State ${\bf \Pi}_{ii} =(2\left \| \mathbf{Q}^i \right \|_2)^{-1}$ if ${{\bf Q}}_i\neq 0$
                \State  $\left ( {\bf B}, {\bf \Lambda }   \right ) = evd\left ( {\bf V}{\bf V}^T \right )$
                \State ${\bf P} = max[ (\mathbf{X}^T\mathbf{X} + \lambda_1\mathbf{G}+\lambda_3 \mathbf{L})^{-1} {\bf{X}}^T \mathbf{U},0]$ 
                \State ${\bf \Xi } = \mathbf{XP}+ \mathbf{K}{{\bf X}}{\bf V}^T$
                  ${\bf H}_{ij} = [{\bf \Xi B}]_{ij}/ ({\bf K}_{ii}{\bf \Lambda}_{jj}+1)$
                \State ${\bf U} = {\bf H}{\bf B}^T$
                \State ${\bf K}_{ii} = \left ( 2\left \| {{\bf X}}_i-{\bf U}_i {\bf X}  \right \|_2 \right )^{-1}$ if ${{\bf X}}_i-{\bf U}_i {\bf X}\neq 0$
            \EndWhile\\
            \Return{$\bf P,Q$}
		\end{algorithmic}
	\end{algorithm}\vsa
\end{center}

\noindent{\bf Overall algorithm, complexity and convergence:} The detailed steps of the UISS are presented in Algorithm~\ref{alg:opt} and follow our update derivations above. We considered both iteration to convergence for each of the sub-problems independently and iterations of one update for each of the variables. The former strategy results in faster convergence for $\bf{W}$ and $\bf{V}$, hence this is the one we present in the Alg.~\ref{alg:opt}. It is important to note that all updates to diagonal matrices (Steps 5, 15, 21) are applied after they are re-initialized to all-zeroes at each iteration.

The complexity of our algorithm is determined by the number of iterations to convergence (in both loops) and the complexity of individual steps in the loops. 
The most costly steps involve the SVD operation in (Step 10) and the matrix inversions (Steps 13 and 17). 
While $svd$ has a super-quadratic complexity in the worst case, efficient implementations enable good scalability in practice. 
The matrix inversions can also be implemented efficiently due to the low-rank updates in each iteration. Note that in Step 13 ${\bf XX}^T$ and in Step 17 ${\bf X}^T{\bf X} + \lambda_3{\bf L}$ are constants and can be inverted once and re-used in subsequent iterations, thus exploiting the sparse structure via the Sherman-Morrison formula for sparse inverse updates: \vsb
\begin{equation}
(\mathbf{A}+\gamma\mathbf{D})^{-1}=\mathbf{A}^{-1}-\frac{\mathbf{A}^{-1}\mathbf{d}\mathbf{d}^T\mathbf{A}^{-1}}{1+\mathbf{d}^T\mathbf{A}^{-1}\mathbf{d}},\vsb
\end{equation}
where $A$ corresponds to the constant matrices in the two steps, $\mathbf{d}=\sqrt{\gamma diag(\mathbf{D})}$ is a column vector and the square root is applied element-wise thus $\gamma\mathbf{D}=d^Td$. 
Therefore, we can reduce the complexity of inversion as $O\left ( mn_{N} \right )$, where $n_{N}$ is the number of non-zero elements in $\bf X$.
The pair $({\bf D},\gamma)$ from the above formula corresponds to $({\bf \Pi},\lambda_2)$ in Step 12 and $({\bf G},\lambda_1)$ in Step 16 respectively. 
\emph{We will publish the implementation of \ourmeth at the camera-ready paper.}

All subproblems in the Alg.~\ref{alg:opt} are solved in an alternating manner and have closed-form solutions, where  $\left \{ {\bf P,Q}, {\bf U}, {\bf V} \right \}$ are optimal solution of Eq.~\ref{objective}. These closed-form solutions guarantee
the overall optimization sequence converges to the primal-dual optimal solutions.
Meanwhile, the constraints in Alg.~\ref{alg:opt} will converge to zero, i.e. $\left \|  \mathbf{P}^{iter+1} - \mathbf{P}^{iter} \right \|_F\rightarrow 0,\left \|  \mathbf{Q}^{iter+1} - \mathbf{Q}^{iter} \right \|_F\rightarrow 0 $. Therefore,  we can obtain the global convergence of \ourmeth because it features both sequence convergence and constraint convergence~\cite{He2002}.

\begin{table}[!t]
\centering
\footnotesize
\begin{tabular}{|l|c|c|c|c|c|}
 \hline
Dataset& $|\mathcal{V}|$ & $|E|$ & $|\mathcal{DS}|$  & Hidden Classes \\
 \hline
 Synthetic    & 100    & 637  & 400   & positive/negative\\
 Bike~\cite{BostonBike}  & 142    & 1,723   & 299   &  weekday/weekend\\
 Embryo~\cite{dutkowski2011protein}&  1,321 &  5,227 & 34  & tissue layers \\
  CCT~\cite{Modeling15-Chen} &  4,665   & 270,571  & 184    & weekday/weekend  \\
 ADNI~\cite{ADNI}    & 6,216  & 683,760 &  173  & AD/NC\\
 Liver~\cite{liver_metastasis}    & 7,383  & 251,916 &  123  & disease/control \\
 \hline
\end {tabular}
\caption{
Summary of evaluation datasets' statistics. }
\label{table:dataset}\vsa\vsb
\end{table}

\section{Experimental Evaluation}
\subsection{Datasets.}

We employ both synthetic and real-world datasets for evaluation and summarize their statistics in Table.~\ref{table:dataset}. 
We synthesize geometric networks by uniformly sampling nodes in a unit square and connecting nodes at distances smaller than a threshold of $\tau=0.2$. We select well-connected ground truth subgraphs and generate balanced set of instances labeled by two (hidden) global states.  Particularly, nodes from the ground truth are assigned a value between $[50,100]$ for positive instances and $[-100,-50]$ for negative ones. Then we add Gaussian noise to all nodes, where ground truth and non-ground-truth nodes have different mean values set to $10$ and $70$ respectively. We also inject outlier instances in the data for some of our experiments.

Nodes in the \emph{Bike}~\cite{BostonBike} dataset are bicycle rental stations in Boston and edges connect stations based on a distance threshold. Nodes' values correspond to the number of check-outs in a day, where we employ the last $299$ days of the data.
We also employ two gene expression datasets: 
\emph{Liver} metastasis~\cite{liver_metastasis} and Embryo~\cite{dutkowski2011protein}. Their network structures are PPI networks~\cite{Dannenfelser2012}, while node features correspond to 
gene expression values with hidden global labels: healthy/normal subjects in Liver and tissue type in Embryo. 
The \emph{ADNI}~\cite{ADNI} dataset contains fMRI resting state measurements for subjects labeled by AD: suffering Alzheimer's disease and NC: healthy normal controls.
The graph structure associates functional links (nodes) with their level of coherence (feature values). Nodes are connected if the corresponding functional links share a brain region. 
\emph{CCT} contains city cellular HTTP traffic data records~\cite{Modeling15-Chen} where nodes are stations, hourly requests are feature values, and node pairs are connected based on a distance threshold. Hidden global labels reflect if the snapshot occurred during workday hours (8am-16pm) or off hours. 


\begin{table}[!t]
\footnotesize
\centering
\begin{tabular}{|l|l|l|l|l|}
\hline
Methods & FS  & IS & Netw. & Description\\ \hline
ALFS~\cite{Li2018alfs}    &        \checkmark           &        \checkmark             &        & Optimal reconstruction \\
\hline
DSLw/o~\cite{Zhang2019dsl}    &         \checkmark           &                    &   \checkmark          & Sparse  self-representation  \\
\hline
UFSOL~\cite{8019357}    &        \checkmark            &                    &        &  Locality preservation \\ \hline
RRSS~\cite{NieRRSS13IJCA}     &                   &         \checkmark            &       &  Subspace learning \\ \hline
{\bf UISS}    &        {\bf \checkmark}           &        {\bf \checkmark}             &   {\bf \checkmark}   &   {\bf Our method} \\ \hline
\end{tabular}
\caption{
Summary of competing techniques. FS: feature selection; IS: instance selection; Network: employs the inter-feature network.}
\label{tbl:baselines}
\end{table}

\subsection{Experimental setup.}

{\noindent \bf Baselines:} All competing techniques are summarized in Tbl.~\ref{tbl:baselines}. We employ several recent feature selection methods to conduct comparison experiments, ALFS~\cite{Li2018alfs},
employs data reconstruction to find the instances and features that minimize the reconstruction error; 
DSLw/o is a variant of the state-of-art supervised discriminative subgraph learning method~\cite{Zhang2019dsl} which we restrict to the unsupervised setting by removing the SVM-like supervision from its objective.
UFSOL~\cite{8019357} 
selects features by preserving the local structure in the data.  
We also employ two recent unsupervised instance selectors: ALFS~\cite{Li2018alfs} which selects instances and features for minimum-reconstruction error and RRSS~\cite{NieRRSS13IJCA} which selects instance based on sparse subspace learning.

{\noindent \bf Metrics:} 
Following the typical experimental setup in unsupervised feature~\cite{8019357} and instance~\cite{Li2018alfs} selection, we first perform these tasks in an unsupervised manner, then reveal the hidden class labels and measure the accuracy of the selected features and/or instances. To allow for fair comparison, we employ the same classifier SVM (linear kernel, $C=1$) for all methods' selections and report average accuracy of $50$ runs.
For datasets with ground-truth features (Liver and Synthetic) and synthetically injected outliers, we also measure the ROC of recovering these ground truth elements respectively. 

\subsection{Experiments on synthetic network data.}
\noindent {\bf Subnetwork selection:} We first evaluate the ability of competing techniques to detect injected ground truth (GT) feature subgraphs in synthetic data. 
The AUC for ground truth subgraph detection with increasing noise variance is presented in Fig.~\ref{fig:syn_noise_fs_roc}. While all methods' performance decreases with increasing noise variance, \ourmeth's ability to recover the injected GT subgraphs consistently dominates that of baselines.
Though DSLw/o uses the network structure for subnetwork selection, similar to \ourmeth, it does not optimize unsupervised discriminative power and is sensitive to noise and outliers. The comparatively lower accuracy of UFSOL is due to the low robustness to noise of its cluster-based feature selection strategy. Similarly ALFS deteriorates quickly with the noise variance due to its preference for high-variance features dominating its Frobenius reconstruction loss. 
\ourmeth is less sensitive to noise due to the $L_{2,1}$ reconstruction loss, the use of the network structure among features and the imposed representative power via orthogonality.

\begin{figure*} [!ht] 
    \centering
    \begin{subfigure}[t]{0.245\textwidth}
        \centering
        \includegraphics[width=\textwidth]{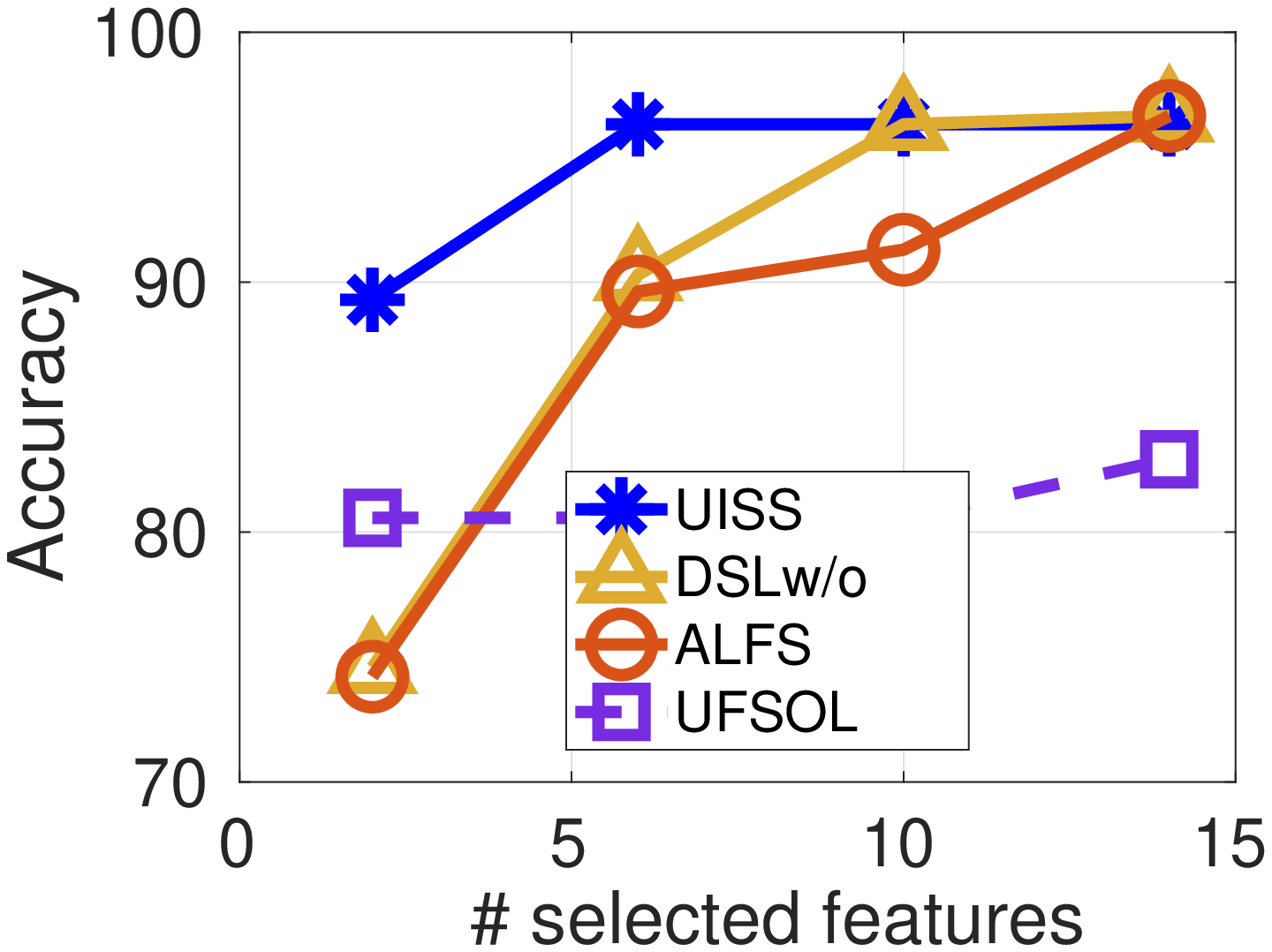}
        \caption{\footnotesize Bike}\label{fig:Boston_accuracy}
    \end{subfigure}
    \begin{subfigure}[t]{0.23\textwidth}
        \centering
        \includegraphics[width=\textwidth]{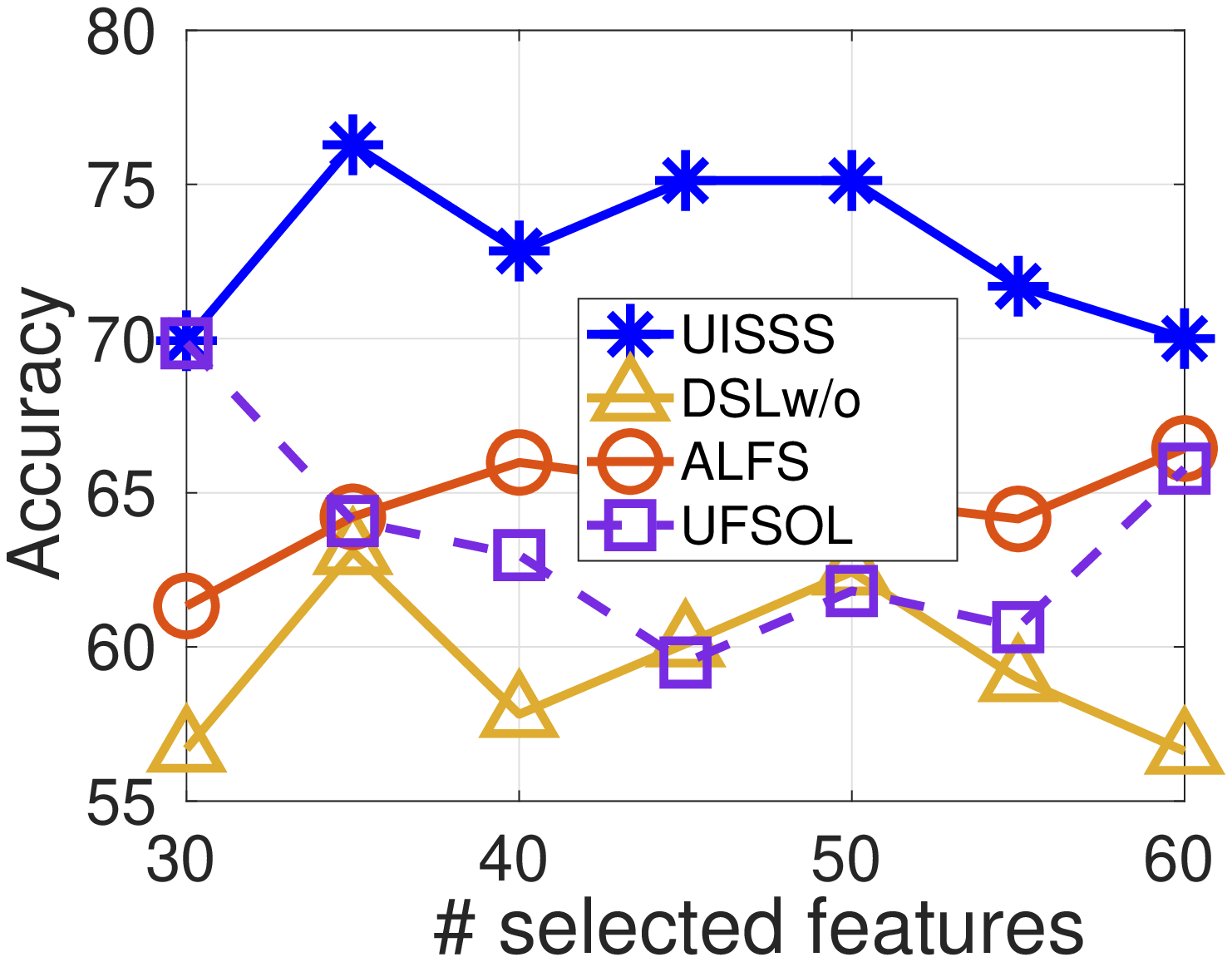}
        \caption{\footnotesize Embryo }\label{fig:Embryo_accuracy}
    \end{subfigure}
           \begin{subfigure}[t]{0.23\textwidth}
        \centering
        \includegraphics[width=\textwidth]{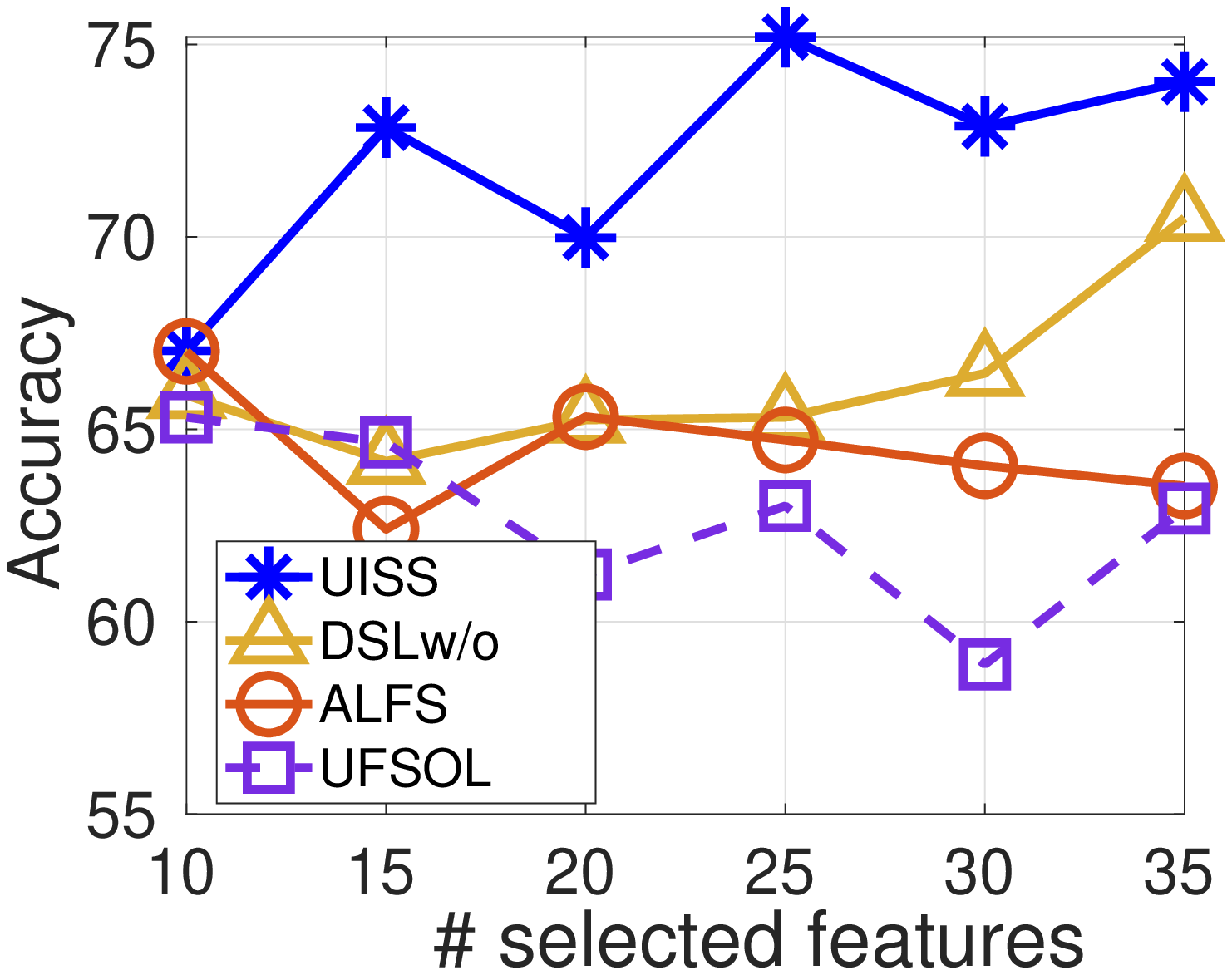}
        \caption{\footnotesize ADNI} \label{fig:Liver_accuracy}
    \end{subfigure} 
    \begin{subfigure}[t]{0.245\textwidth}
        \centering
        \includegraphics[width=\textwidth]{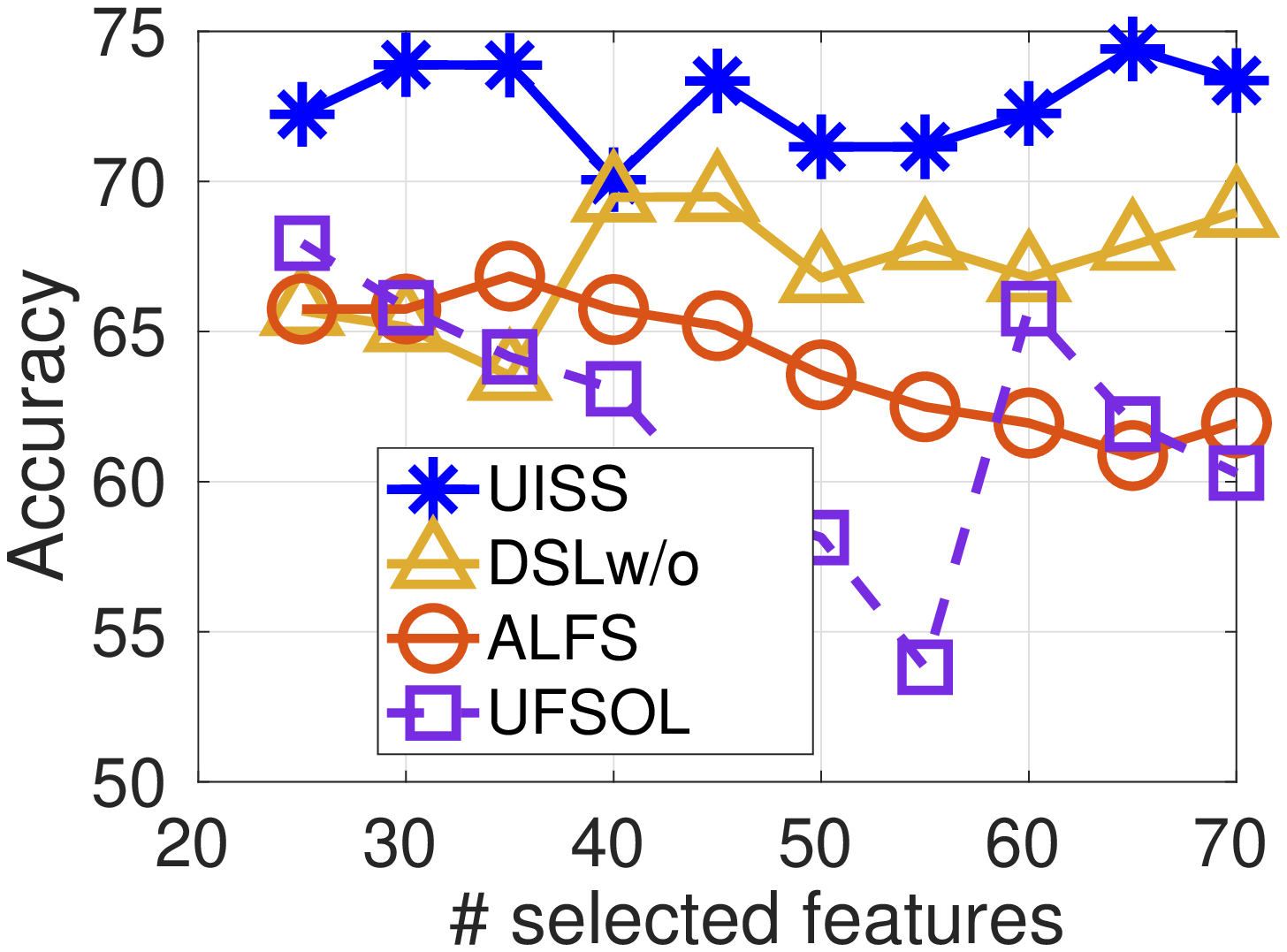}
        \caption{\footnotesize CCT} \label{fig:Liver_accuracy}
    \end{subfigure}

    \caption{
    Comparison of classification accuracy on real world datasets with increasing number of selected features.}
    \label{fig:whole_fs_acc} \vsa\vsb
\end{figure*}

We also study the robustness of subnetwork selection for increasing the number of outlier instances and provide comparison in Fig.~\ref{fig:syn_outlier_fs_roc}. \ourmeth dominates and degrades more gracefully with the number of outlier compared to alternatives. This robustness can be attributed to the combination of orthogonality of selected instances for self-representation and the $L_{2,1}$ reconstruction cost, collectively discarding outliers as unable to represent other instances and incurring high reconstruction error.
All baseline methods' AUC suffers at all levels of outliers since they treats all instances as cluster members and do not explicitly model the existence of outliers. 

\noindent {\bf Instance selection:} Next we evaluate the utility of competing techniques as discriminative instance selectors (Fig.~\ref{fig:syn_noise_is_acc}). \emph{Can selected instances inform accurate classifiers if global labels are added post-selection?} Among the instance selector baselines, ALFS can also perform feature selection while RRSS selects representative instances based on all features. Note that while ALFS and \ourmeth internally re-weigh features top select instances, the final classifiers we train in this experiment are based on all features in selected instances to enable a fair comparison with RRSS.

When all training instances are available, the accuracy of competitors is the same as they employ the same SVM classifier and all features. As we constrain the methods to select smaller number of instances for labeling, the baselines' performance degrades, while that of \ourmeth increases slightly and remains above the all-instance accuracy even when we select only $10\%$ of the total instances (Fig.~\ref{fig:syn_noise_is_acc}). The advantage of our model stems from the dependence of the instance selector $\bf Q$ on the feature selector $\bf P$ which enforces smoothness in the network structure $\mathcal{S}$ as well as the imposed orthogonality on the selected representative instances.
In other words, while we do not explicitly exclude features, \ourmeth treats the available features deferentially which in turn informs better instance selection. This advantage is to some extent noticeable for ALFS as well, although its inability to consider the network structure among features leads to significantly degraded accuracy for decreasing number of selected instances.

\noindent{\bf Outlier detection.} We also employ instance selectors to detect outlier instances injected in the dataset (Fig.~\ref{fig:syn_outlier_is_acc}). We invert the instance selection order reported by each method and plot the ROC for predicting outlier instances (the dataset is balanced, i.e. the same number of outliers and non-outlier instances). 
\ourmeth is better at detecting outliers than ALFS, since the latter employ Frobenius reconstruction strategies, thus forcing outliers' inclusion among important instances.
Although RRSS employs an $L_{2,1}$-norm fitting function, it does not to consider the structure among features when informing the instance selection. Our choice of $L_{2,1}$ norm for reconstruction and the orthogonality regularization on the instance association matrix allows \ourmeth to ``spot'' outlier instances better. 

\begin{figure}[!t]
    \footnotesize
    \centering
    \centering
    \begin{subfigure}[t]{0.23\textwidth}
         \centering
         \includegraphics[width=\textwidth]{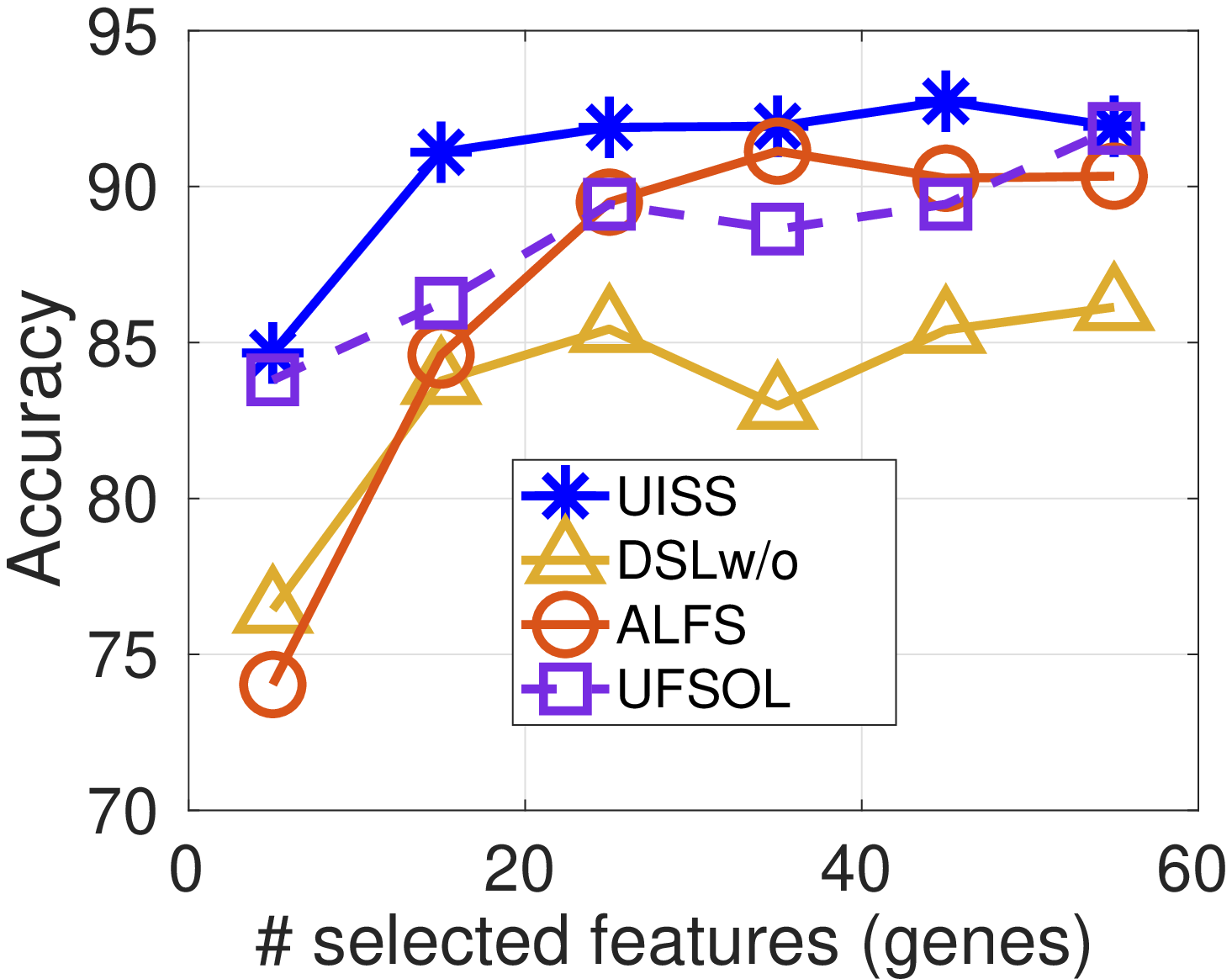}
         \caption{\footnotesize Classification accuracy }\label{fig:liver_acc}
     \end{subfigure}
    \begin{subfigure}[t]{0.24\textwidth}
        \centering
\includegraphics[width=\textwidth]{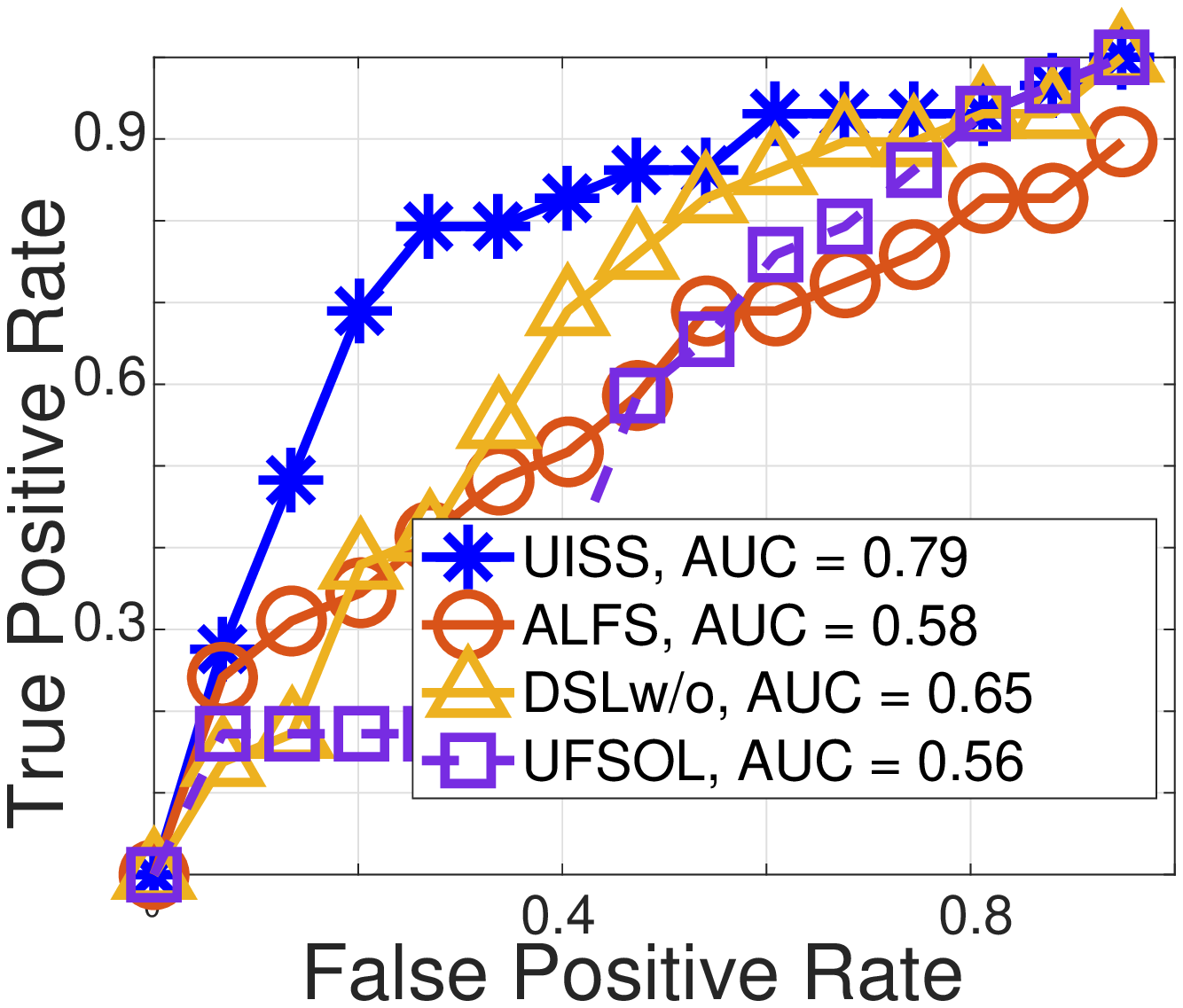}
        \caption{\footnotesize Ground truth features}\label{fig:liver_roc}
    \end{subfigure}
\caption{
Classification accuracy (a) and ROC for selecting ground truth genes (b) in the Liver dataset.}
 \label{fig:Liver_ACC_GT_ROC}
\end{figure} 

\begin{figure*} [!htb] 
    \centering
    \begin{subfigure}[t]{0.22\textwidth}
        \centering
        \includegraphics[width=\textwidth]{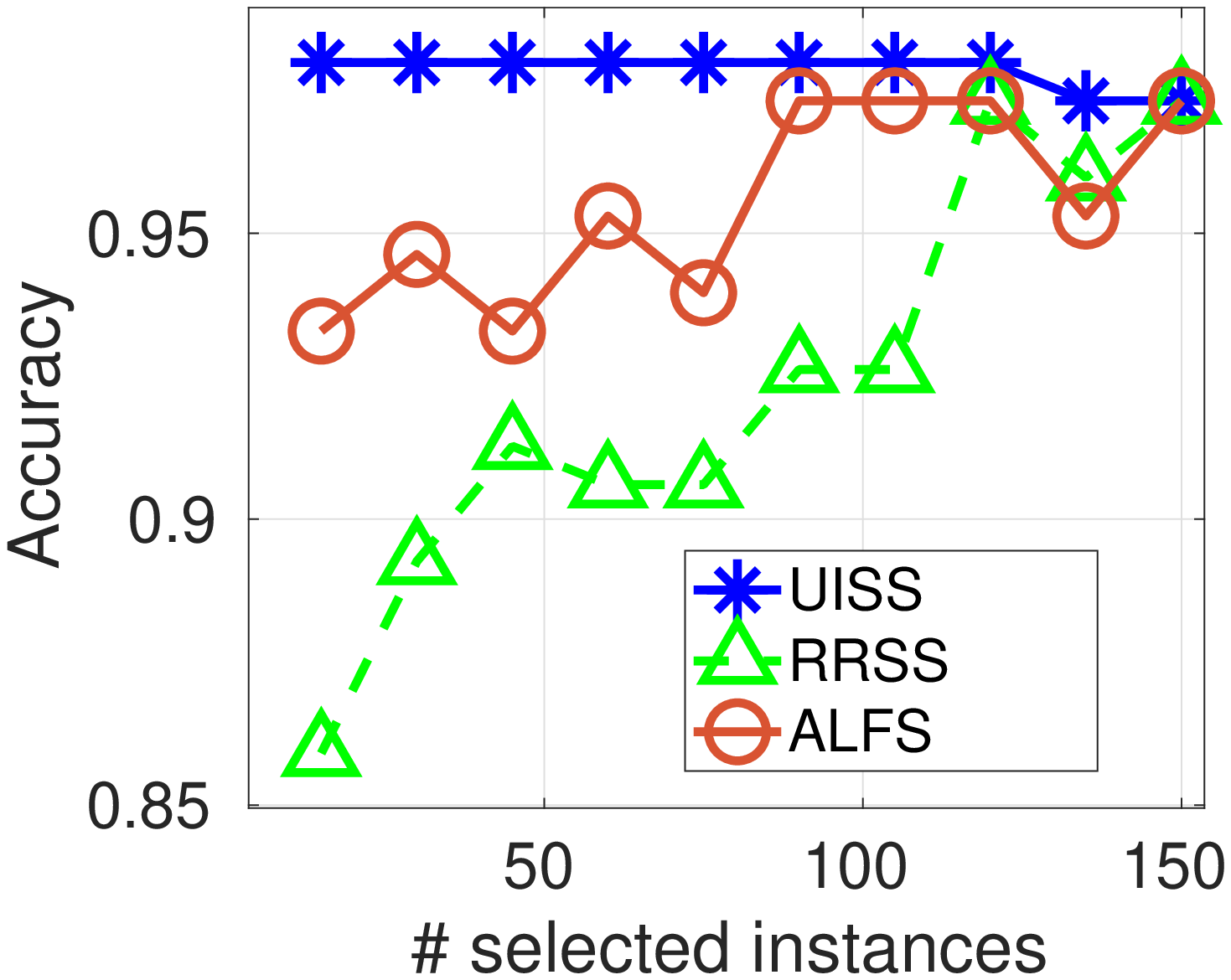}
        \caption{\footnotesize Bike}\label{fig:Boston_is_accuracy}
    \end{subfigure}
    \begin{subfigure}[t]{0.235\textwidth}
        \centering
        \includegraphics[width=\textwidth]{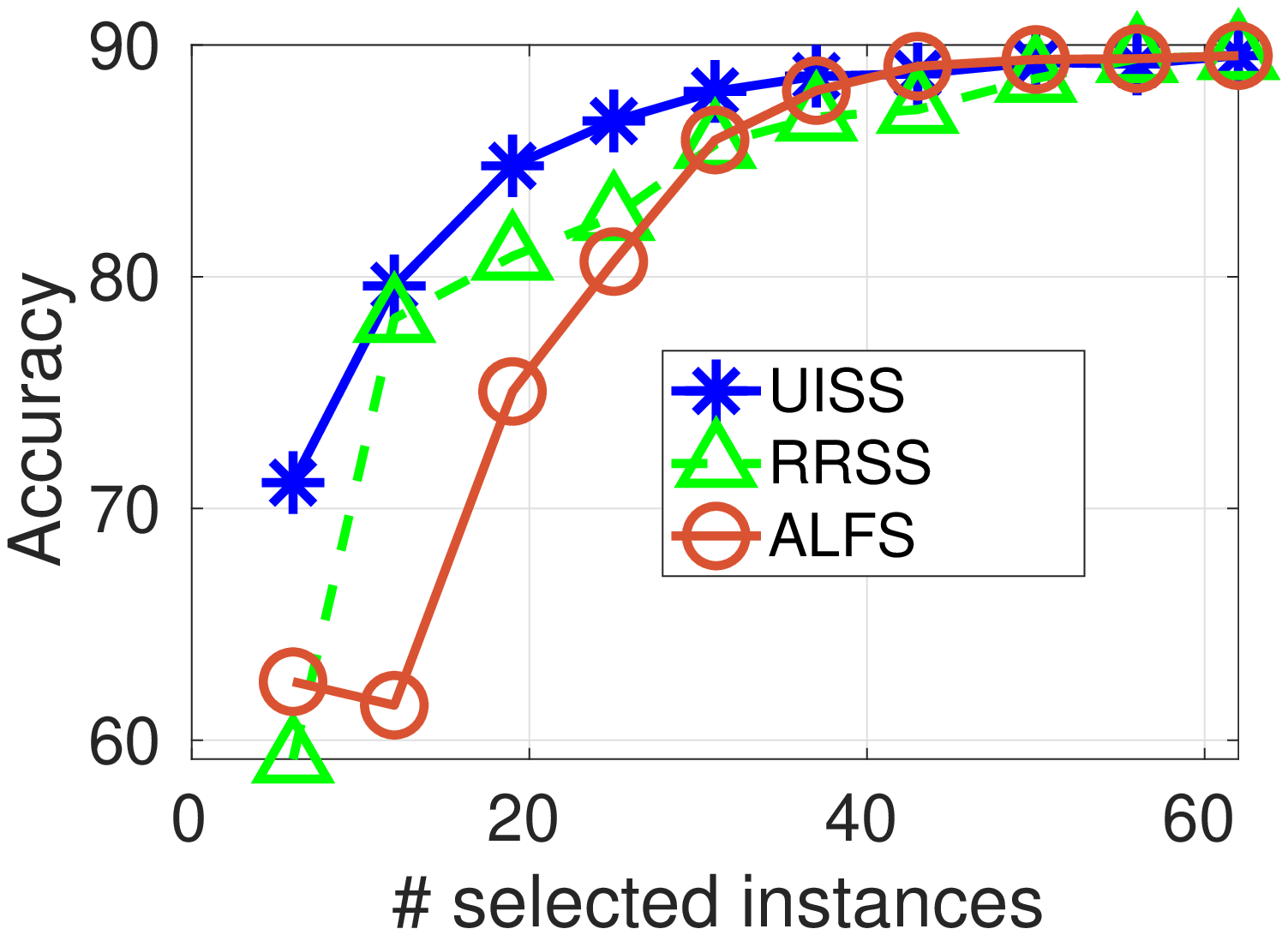}
        \caption{\footnotesize Liver}\label{fig:embryo_is_accuracy}
    \end{subfigure}
    \begin{subfigure}[t]{0.23\textwidth}
        \centering
        \includegraphics[width=\textwidth]{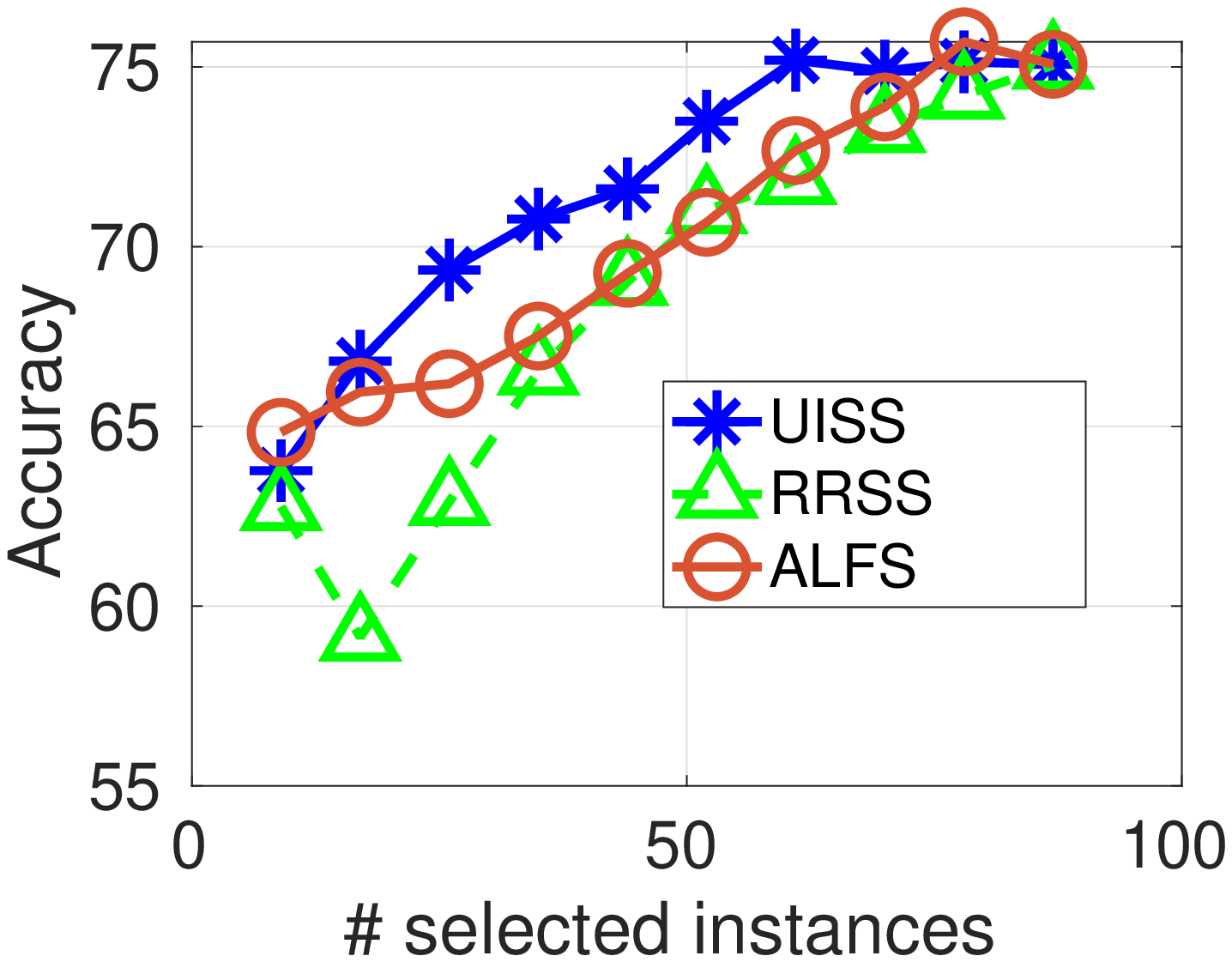}
        \caption{\footnotesize ADNI}\label{fig:embryo_outlier_roc}
    \end{subfigure}
        \begin{subfigure}[t]{0.24\textwidth}
        \centering
        \includegraphics[width=\textwidth]{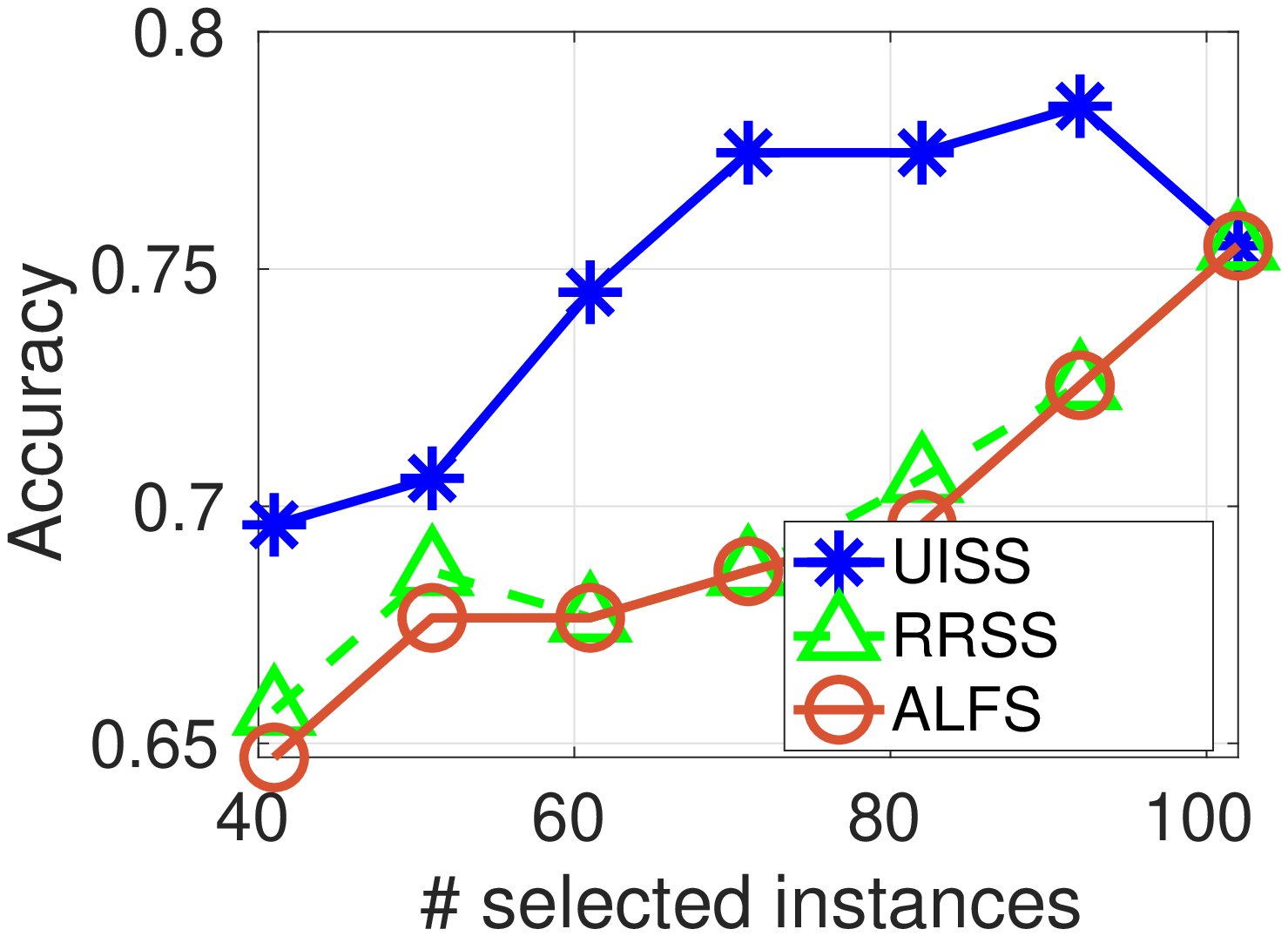}
        \caption{\footnotesize CCT}\label{fig:Boston_outlier_roc}
    \end{subfigure}
    \caption{
    Comparison of classification accuracy with increasing number of selected instances on real world datasets.
    }
    \label{fig:whole_instance_sel}
\end{figure*}

\begin{figure*}[!ht]
    \footnotesize
    \centering
    \centering
    \begin{subfigure}[t]{0.21\textwidth}
         \centering
         \includegraphics[width=\textwidth]{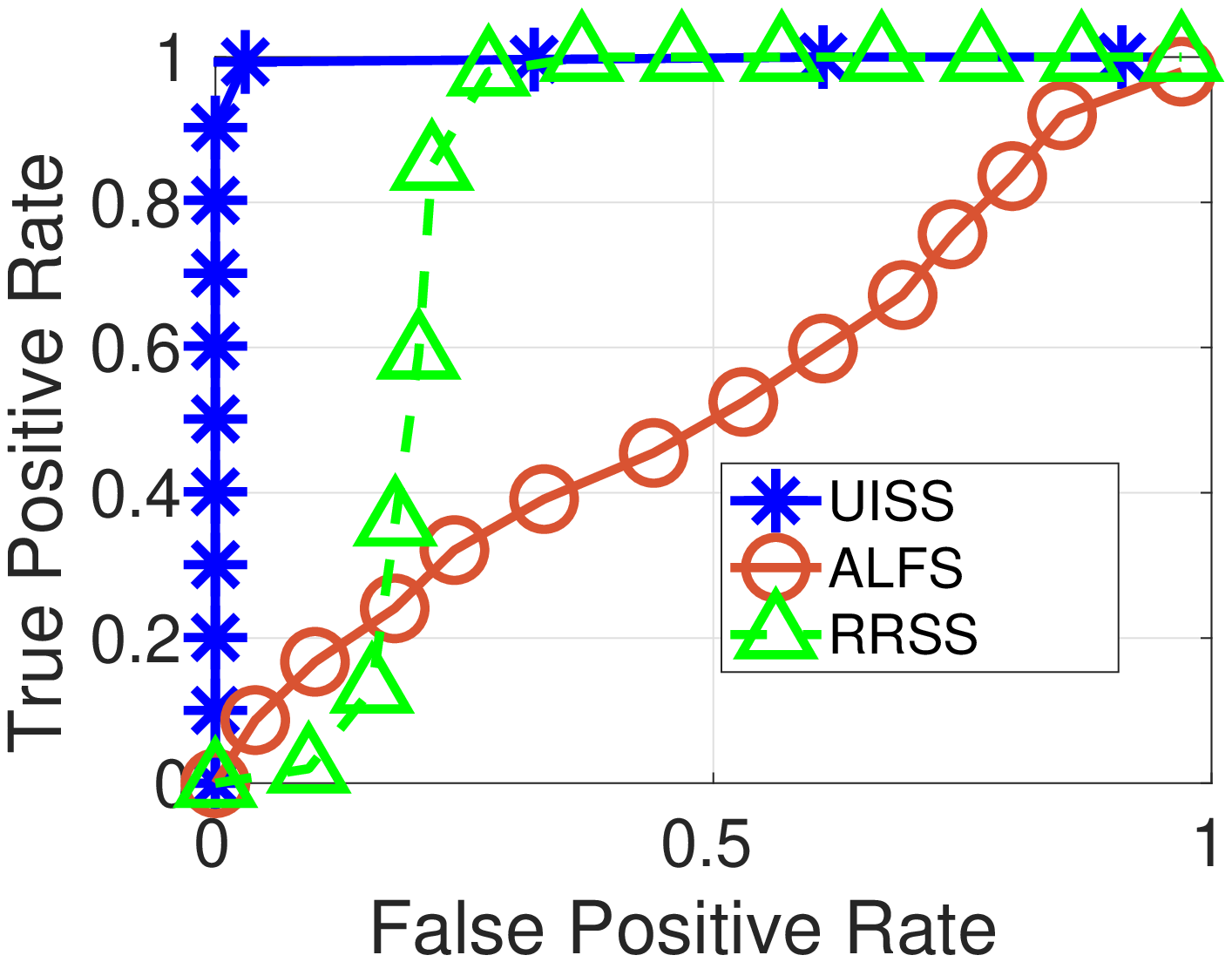}
         \caption{\footnotesize Outliers (Bike)}\label{fig:outlier_boston}
     \end{subfigure}\vsa
    \begin{subfigure}[t]{0.20\textwidth}
        \centering
\includegraphics[width=\textwidth]{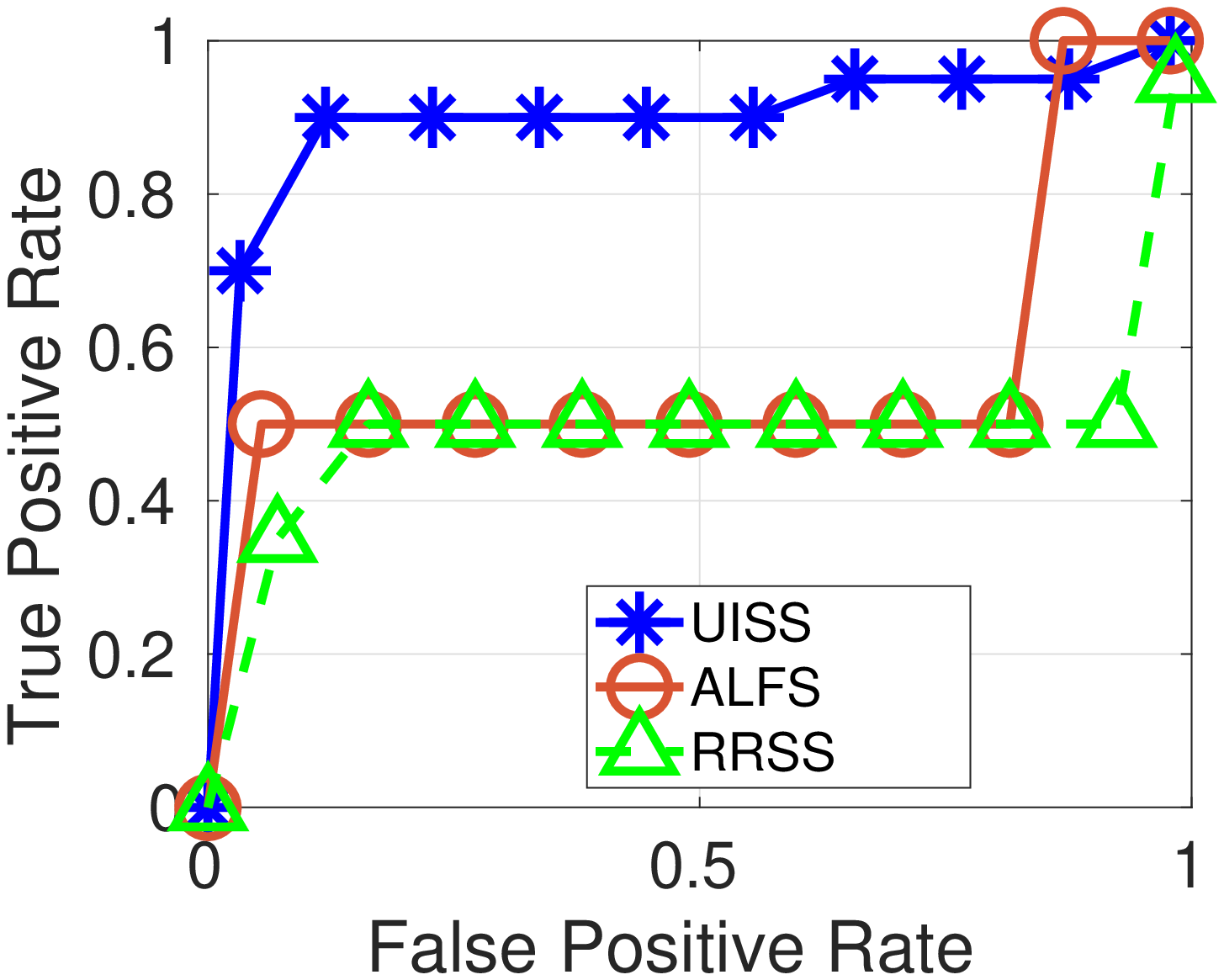}
        \caption{\footnotesize
        Outliers (CCT)}\label{fig:outlier_cct}
    \end{subfigure}
    \begin{subfigure}[t]{0.23\textwidth}
        \centering
        \includegraphics[width=\textwidth]{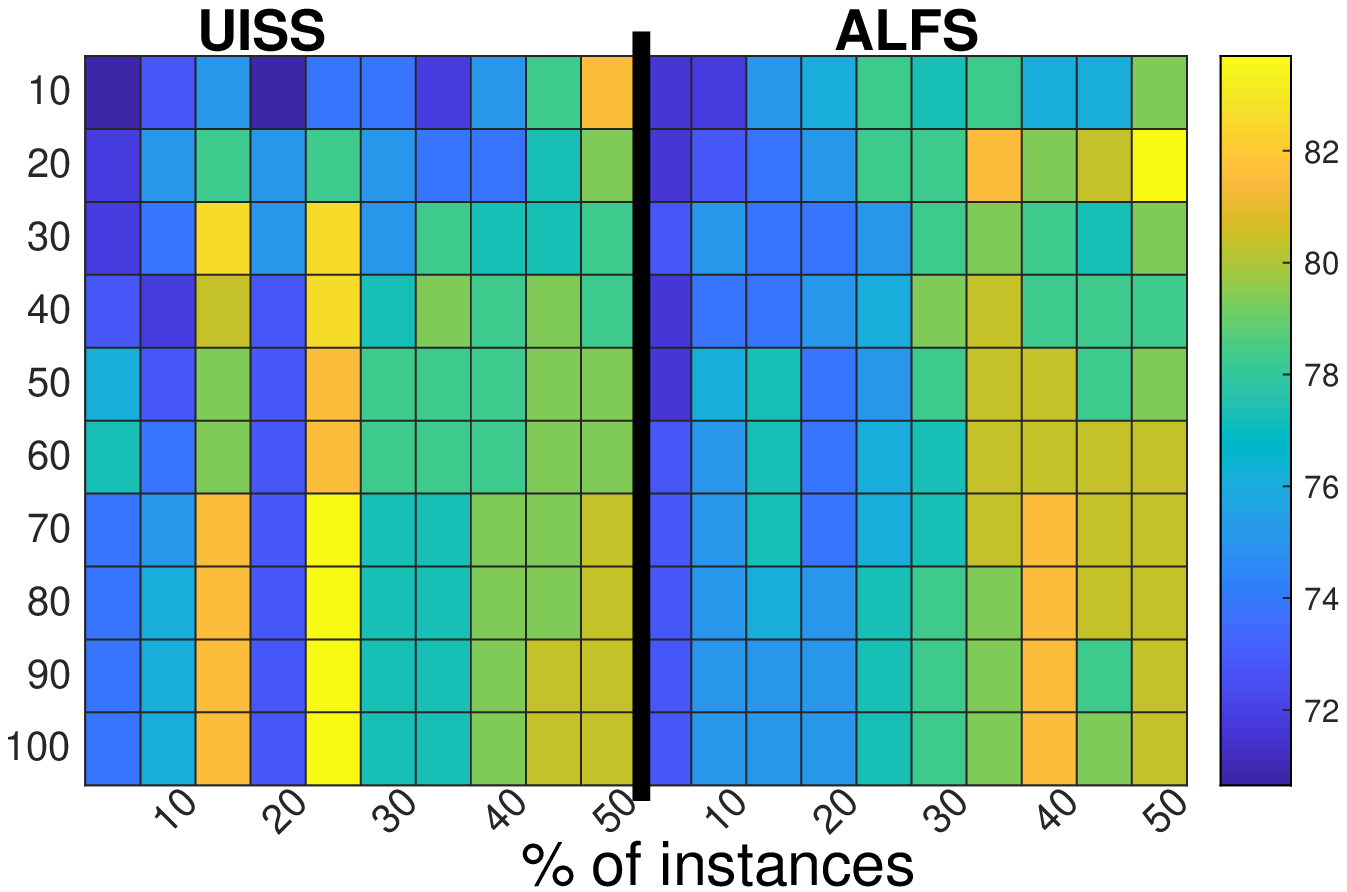}
        \caption{\footnotesize 
        Joint FS and IS }\label{fig:FSIS_ACC}
    \end{subfigure}
    \begin{subfigure}[t]{0.175\textwidth}
         \centering
         \includegraphics[trim={1cm 0 0 1cm},clip,width=\textwidth]{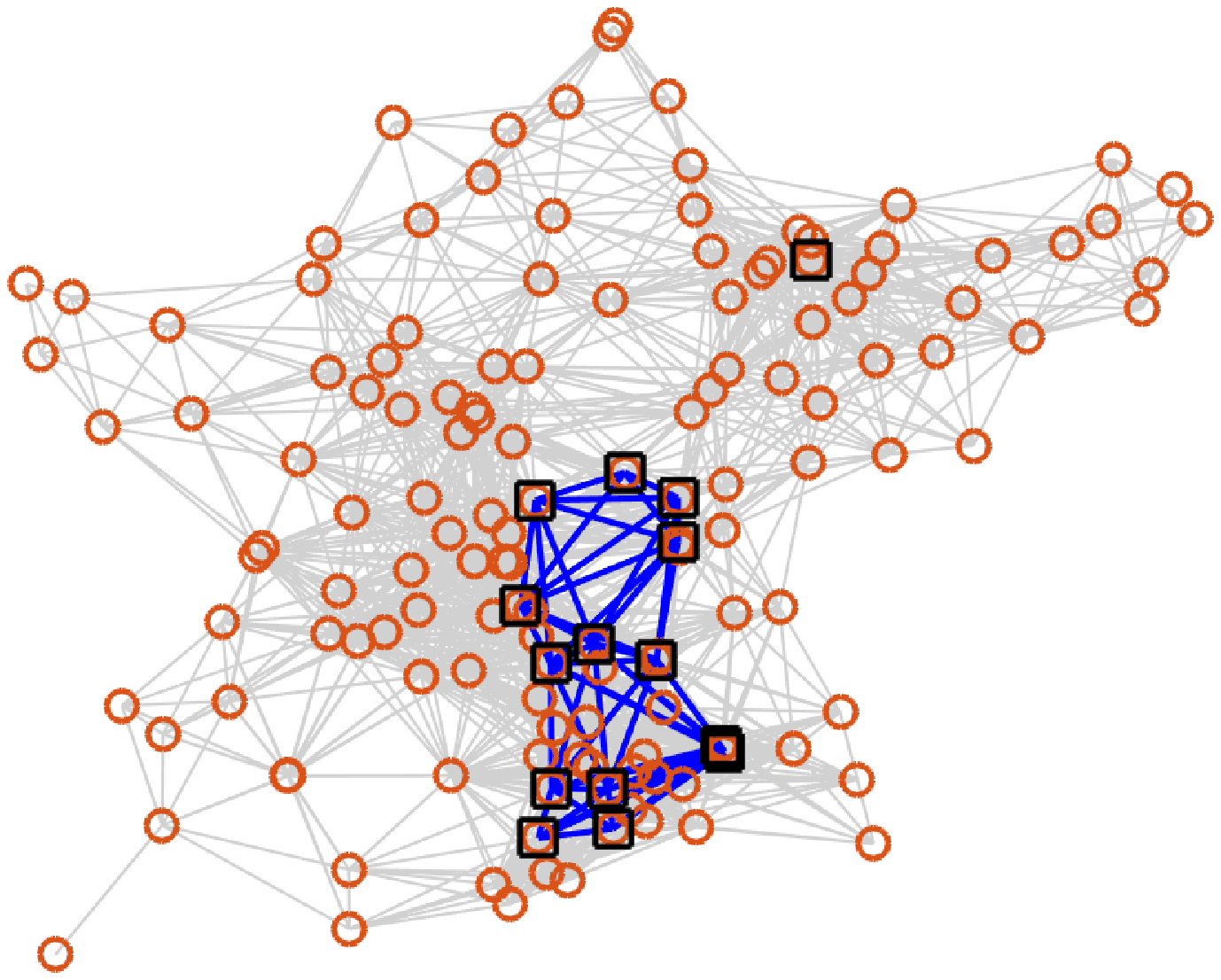}
         \caption{\footnotesize
         UISS }\label{fig:vis_boston}
     \end{subfigure}\hspace{-0.5cm}
    \begin{subfigure}[t]{0.175\textwidth}
        \centering
\includegraphics[trim={1cm 0 0 1cm},clip,width=\textwidth]{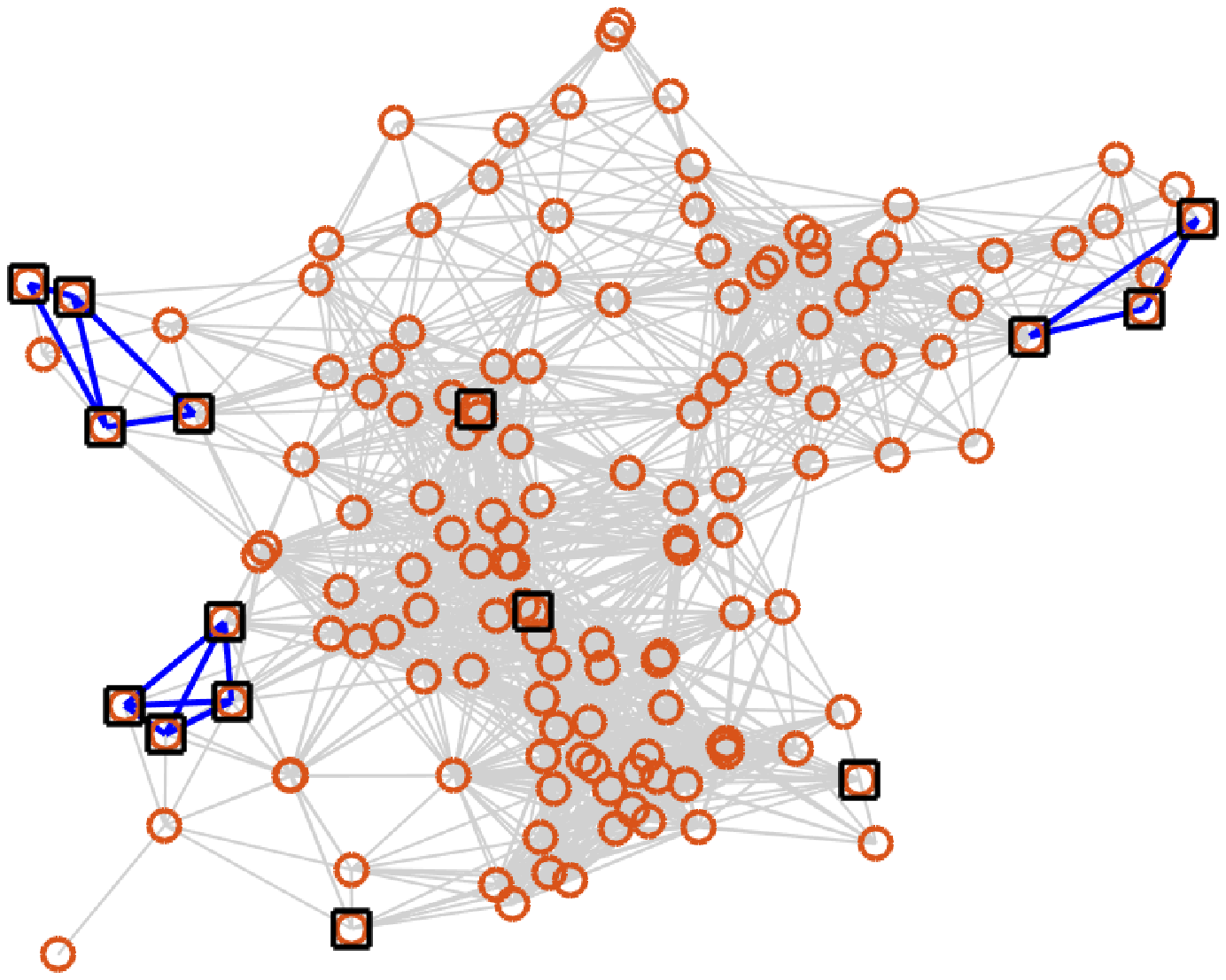}
        \caption{\footnotesize
        UFSOL }\label{fig:vis-udfs_boston}
    \end{subfigure}
\caption{
(\subref{fig:outlier_boston}),(\subref{fig:outlier_cct}): Outlier detection on real-world datasets. (\subref{fig:FSIS_ACC}) 
Accuracy comparison of joint unsupervised feature and instance selection between \ourmeth (left) and ALFS (right) on the CCT dataset. The number of selected features varies on the vertical axis. \ourmeth achieves over $82\%$ accuracy with as few as $30$ features and $30$ instances, while ALFS requires all instances to attain the same performances. (\subref{fig:vis_boston}),(\subref{fig:vis-udfs_boston}): Visualization of the subgraphs selected by UISS and UFSOL on the Boston Bike dataset.}
 \label{fig:outlier_detection} \vsa\vsb
\end{figure*}
\subsection{Experiments on real-world network data}
\noindent{\bf Subnetwork selection: } Next we vary the number of features in the selected subnetwork 
and then add the hidden labels to quantify the cross-validation accuracy (based on selected features in all instances) in the network datasets (Fig.~\ref{fig:whole_fs_acc}). \ourmeth's accuracy dominates all baselines on all datasets and its accuracy is relatively stable around its optimal value. A second pattern that stands out is that there is no clear second best baseline. 
While DSLw/o dominated other baselines in identifying GT features in synthetic (Figs.~\ref{fig:syn_noise_fs_roc}), its ability to select predictive features on real datasets does not consistently dominate non-network baselines: better on CCT, ADNI and Bike, but worse on Embryo where features are binary. The best improvement of \ourmeth over baselines in these datasets is about $10\%$ of accuracy and it is retained over different number of selected features.

In addition, we quantify the subgraph selection quality for the Liver dataset that has partial ground truth genes associated
with the global disease state labels and reported in~\cite{DIPS} (Figs.~\ref{fig:Liver_ACC_GT_ROC}). 
We observe that the methods which consider the PPI network structure among features (genes), namely \ourmeth and DSLw/o, are better at recovering known genes associated with the disease than alternative which do not consider this structure. In particular the AUC of \ourmeth and DSLw/o are $0.79$ and $0.65$ respectively, while those of non-network alternatives are close to random (Fig.~\ref{fig:liver_roc}). Without considering the network structure, UFSOL and ALFS tend to select isolated nodes rather than a connected subgraph and those are not part of potential target pathways which form connected subgraphs in the PPI network. UFSOL achieves the second best performance after \ourmeth in terms of classification accuracy (\ref{fig:liver_acc}), however, its AUC for ground truth gene detection is the worst among competitors (Fig.~\ref{fig:liver_roc}). It is important to note that the set of ground truth genes as discussed in~\cite{DIPS} is far from complete, i.e. there are potentially more genes associated with the disease, and thus the competing techniques could be employed to detect more target genes and pathways of biological significance.

\noindent{\bf Instance selection:}
Similar to synthetic data, we also compare instance selectors on classification accuracy for increasing number of informative instances selected in an unsupervised manner (Fig.~\ref{fig:whole_instance_sel}).
The observed behavior in real-world dataset is similar to that in synthetic data. Namely, \ourmeth dominates alternatives for small number of selected instances and it achieves its optimal accuracy without using all available instances. This is a very promising result for applications in which acquiring annotations is expensive or time consuming, since employing \ourmeth would enable the creation of accurate classifiers with minimal number of instances.

\noindent{\bf Outlier detection:}
We also evaluate the ability of competing techniques to detect synthetically injected outliers in real-world datasets in Fig.~\ref{fig:outlier_detection}. As in synthetic, we invert the scores for instances to include, i.e. the lowest-ranking instances are deemed highest-scoring outliers. Injected outlier instances have random feature values with similar mean value to instances in the respective real-world dataset. \ourmeth achieves close-to-optimal performance on the Bike and CCT datasets, while ALFS behaves close to random due to its sensitivity to outliers. RRSS is the second best method on the Bike dataset as it explicitly models outliers for instance selection. The behavior of competitors on CCT is also close to random, although in this dataset their initial FPR growth is steeper and later flattens due to ranking some outliers among the most important instances. 

\subsection{Joint feature and instance selection.}
\label{sec:joint-fs-is}
To enable a fair comparison with non-network baselines in our feature and instance selection experiments from the previous subsections, we sub-selected one of those dimensions and used the other one fully. The performance of \ourmeth improves both when some number of features and a subset of all available instances are excluded. The optimal subsets in both dimensions are likely inter-independent, and hence, in what follows, we study this dependence by an exhaustive sub-selection of both instances and features in the CCT dataset. Note that this dataset is very high dimensional and learning reliable predictive models on it is likely to benefit from both feature and instance selection. The only baseline that performs feature and instance selection jointly is ALFS, and hence, we focus on comparison between \ourmeth and ALFS in this experiment. 

We vary the number of selected features from $10\%$ to $100\%$ (top to bottom in Fig.~\ref{fig:FSIS_ACC}) and the number of instances between $5\%$ and $50\%$ (left to right in Fig.~\ref{fig:FSIS_ACC}) in the CCT dataset and present the cross-validation accuracy after revealing the hidden labels for the selected feature subsets by the methods (Fig.~\ref{fig:FSIS_ACC}). \ourmeth achieves its optimal accuracy ($81\%$) with as few as $30\%$ of the features and $30\%$ of the instances. To get to a similar performance ALFS employs $100\%$ of the instances and $20\%$ of the features, i.e. three times the number of instances and $10\%$ less features. Overall, both methods benefit from the joint selection of features and instances, however, \ourmeth utilizes the network structure and further employs regularization making it more robust to outliers. 



\subsection{Performance on non-network data.}
\begin{figure} [!t]
    \centering
    \begin{subfigure}[t]{0.2\textwidth}
         \centering
         \includegraphics[width=\textwidth]{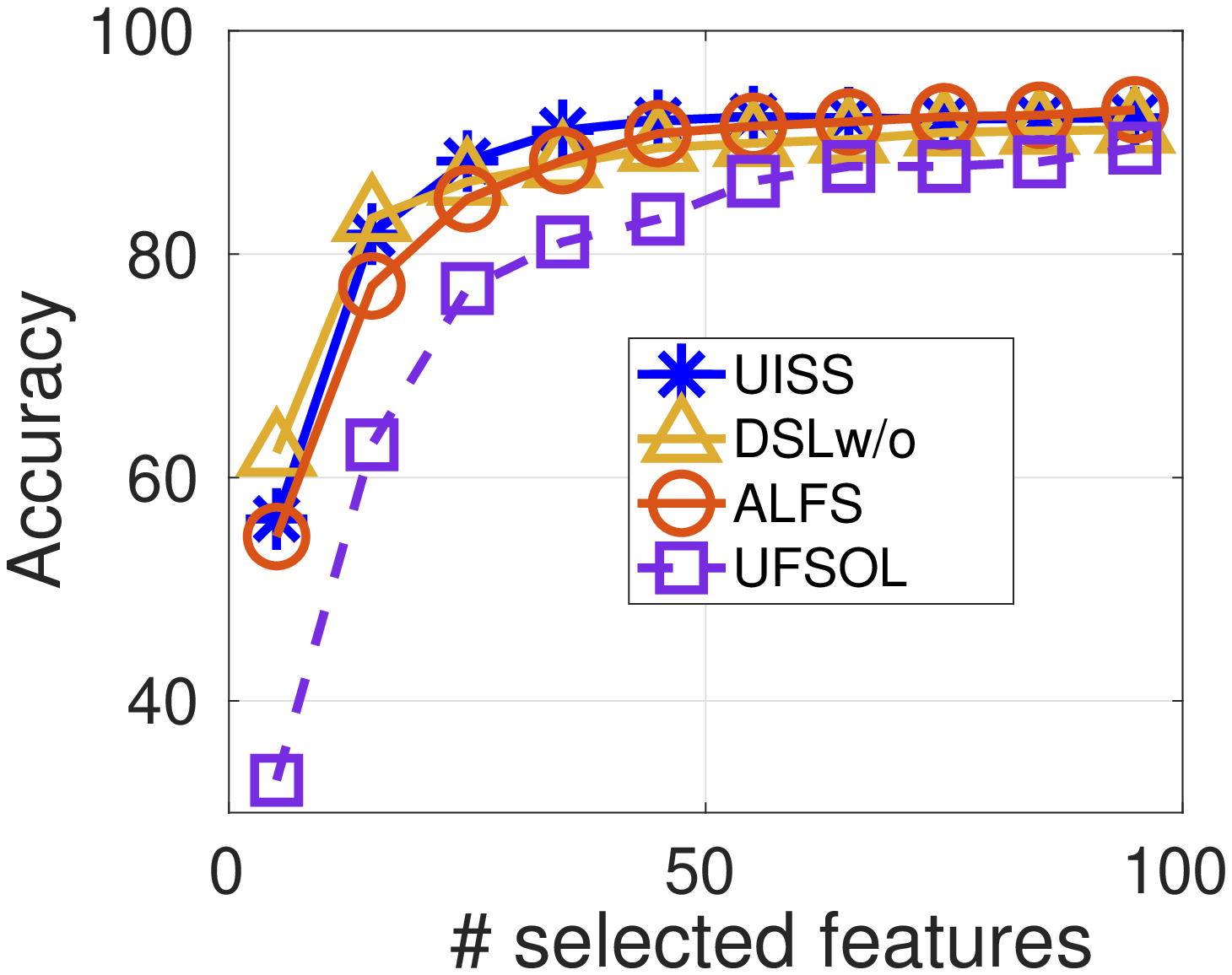}
         \caption{\footnotesize Feature selection}\label{fig:usps_fs}
     \end{subfigure}
    \begin{subfigure}[t]{0.207\textwidth}
        \centering
        \includegraphics[width=\textwidth]{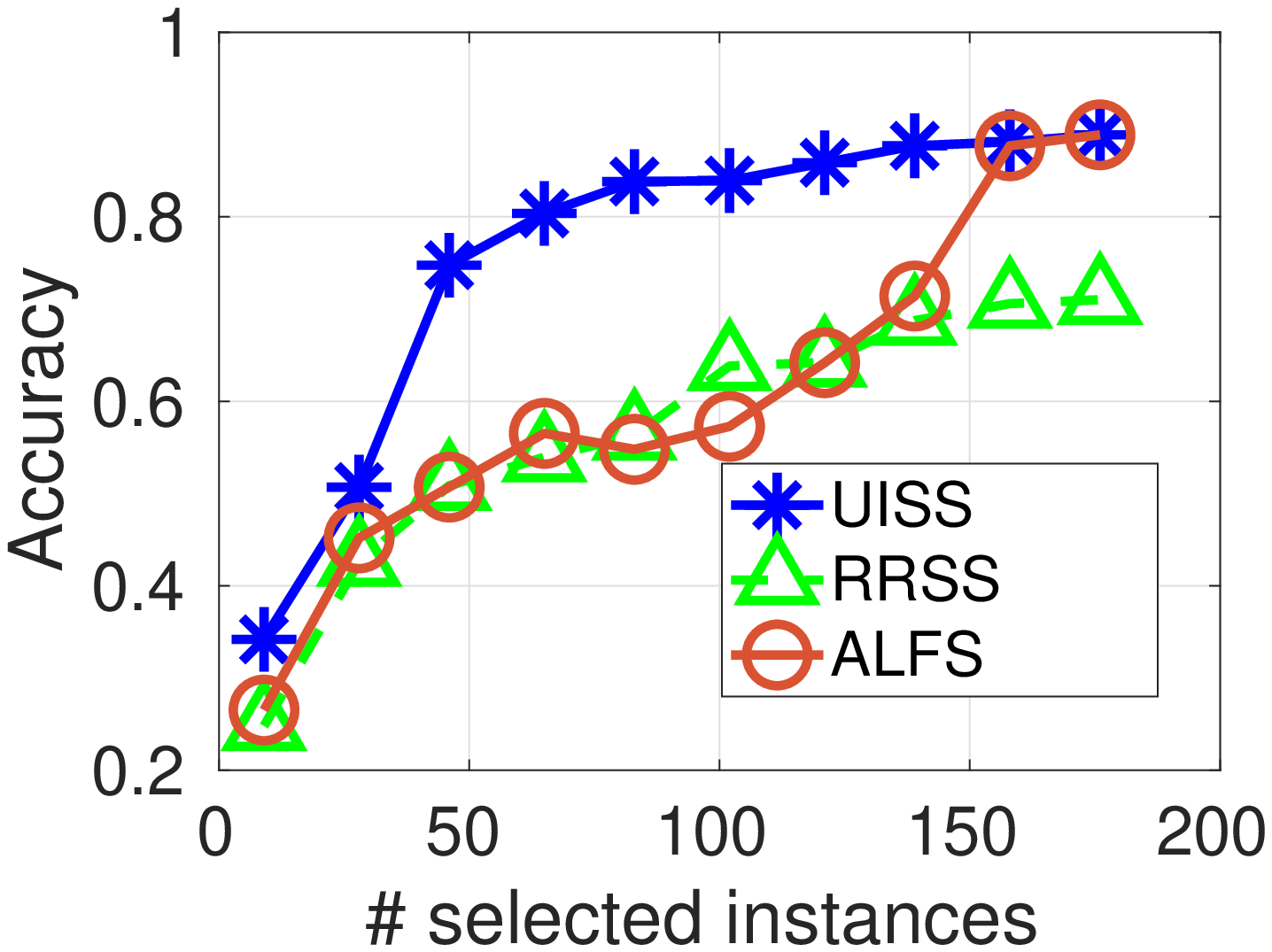}
        \caption{\footnotesize Instance selection}\label{fig:usps_is}
    \end{subfigure}
    \caption{
    Feature and instance selection accuracy on a non-network dataset  USPS~\cite{Belhumeur97eigenfacesvs.}.}
    \label{fig:usps}\vsa\vsb
\end{figure}

\ourmeth is designed specifically as an unsupervised feature and instance selector for network data. However, one important question about its performance is whether its advantage is only due to the availability of network structure among the data features. To test this, we turn to a common image (``non-network'') dataset 
USPS~\cite{Belhumeur97eigenfacesvs.}.
This dataset was previously adopted for both feature and instance selection by some of our baselines and also more broadly in the machine learning literature. We compare a non-network version of \ourmeth ($\lambda_3=0$) to all baselines for feature and instance selection cross-validation accuracy (Fig.~\ref{fig:usps}), following the same protocol as the one adopted for network data. 
Interestingly, when employed for instance selection, \ourmeth once again dominates all baselines when limiting the number of instances employed to train a classifier (Fig.~\ref{fig:usps_is}).
The feature selection experiment renders our method still remains competitive with
state-of-art feature selection methods
when they are restricted to use between $1\%$ and $10\%$ of the available features (Fig.~\ref{fig:usps_fs}).
While these experiments show promise for the generality of \ourmeth beyond network data, investigation of more non-network datasets is necessary to confirm its utility in such settings. Extensive non-network experiments are beyond the scope and space constraints of the current manuscript and we plan to include such analysis in an extended journal version. 


\subsection{UISS at work: salivary gland organoids and bike sharing.}

\begin{figure}[!t]
    \footnotesize
         \includegraphics[trim={5cm 2.5cm 9cm 0cm},clip,width=0.5\textwidth]{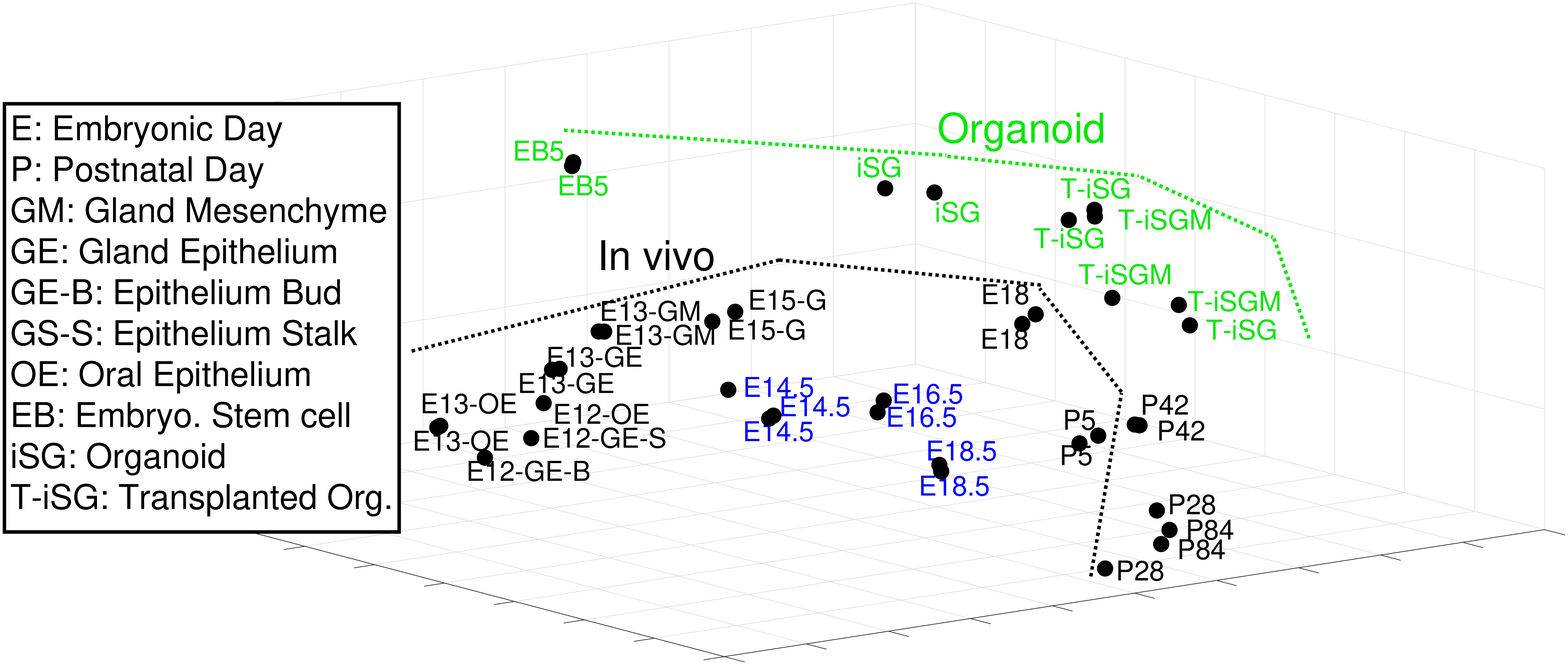}
\caption{
UISS employed to analyze networked RNA-seq samples from salivary gland development. Samples are annotated post-analysis with (i) development time (E12-P84), and (ii) sample types GM/GE/EB/OE/(T)iSG. The embedding learned by UISS reconstructs the developmental timeline and isolates expected stem cell outliers (EB) and differentiates between organoid (iSG) and in vivo development.  ($k=20,\lambda_1=0.1, \lambda_2=0.1, \lambda_3=10^3$).}
 \label{fig:organoids}
\end{figure}

\noindent{\bf Organoids:} 
We next evaluate the ability of UISS to elucidate the organization of RNA sequencing (RNA-seq) networked samples combined from two different studies of mouse salivary gland development~\cite{ tanaka2018generation}. 
RNA-seq quantifies the levels of RNAs present in a tissue sample and the Tanaka study~\cite{tanaka2018generation} employed RNA-seq to characterize the gland organoids derived from embryonic stem cells which are grown ex vivo and engineered to mimic in vivo organ development. 
Organoids offer rapid disease modeling with applications to infectious diseases from Zika to SARS-CoV-V2.
This is a fitting use case for UISS, since while fusing multiple unlabeled, high-dimensional and noisy datasets may enable important new insights, it also warrants examination for outliers as well as artifacts of varying experimental protocols. 
The organoid data set  includes 38 instances: 19 time points (some with replicates), spanning from embryonic day 12 (E12) to 12 weeks into adulthood (P84), as well as different cell types (GE/GM/OE) and manipulations (iSG/T-iSG) labeled in Fig.~\ref{fig:organoids}. 
We employed the fold changes in RNA-seq counts as input features after a regularized logarithm transformation~\cite{love2014moderated}. 

We apply UISS on the data and obtain a 3D embedding of the instances by employing PCA on the learned data projection $\bf XP$ presented in Fig.~\ref{fig:organoids}. It is important to note that the labels in the figure were not used by UISS, i.e. the analysis is fully unsupervised. Embryonic stem cell derived embryoid bodies (EBs) are pluripotent and able to become all known cell types. Not surprisingly, UISS renders them based on their RNA-seq as outliers compared to other samples. After the EBs are induced to differentiate into salivary glands (iSG) in vitro their transcriptional landscape changes to become more similar to developing glands. Using the time course labels (E12 through P84), we can place experimental data in frame with the known developmental (in vivo) progression (black dotted line). Organoids (iSG) are transcriptionally similar to an early salivary gland differentiation stage around E15-16, however, once upon transplantation in vivo (samples T-iSG) the organoid overcome a developmental wall furthering their maturation. The T-iSG transcriptomes should be comparable to postnatal salivary glands from five days after birth (P5) to 84 days after birth (P84) from the Gluck study~\cite{gluck2016rna}.

Interestingly, we detect a wider range of variability in maturation of the transplanted samples, which could represent either variability in the transplanted organoids or differential responses of the individual organoids to the in vivo environment. Although the representation by UISS cannot readily explain the variable gene expression by the organoids, this example of staging organoids illustrates its ability to map relationships between biological datasets. As RNA-seq is broadly used in biomedical sciences, UISS will have broader applicability in comparison of biological datasets. It is important to note that, PCA on the UISS' embedding of the data $\bf XP$ retains over $99\%$ of the variance in the first component, while PCA on the raw data $X$ requires more than $12$ components for $99\%$ variance retention.  

\noindent{\bf Bike sharing:} We also investigate the interpretability of the feature selection by UISS in the Bike dataset. Figs.~\ref{fig:vis_boston},~\ref{fig:vis-udfs_boston} visualize the selected subgraphs (highlighted nodes and edges) by \ourmeth and UFSOL---its closest competitor in terms of accuracy when employing at least $6$ features (see Fig.~\ref{fig:Boston_accuracy}).
As expected, UISS selects a mostly connected feature subgraph due to the network smoothness regularization. This subgraph corresponds to a locality of bike rental stations in downtown Boston which can differentiate between weekday and weekend traffic (hidden classes in this dataset). 
UFSOL selects features which form multiple connected components which are less obvious to interpret and more importantly result in less predictive classifiers (as demonstrated in Fig.~\ref{fig:Boston_accuracy}).

\subsection{Parameter sensitivity and running time}
We evaluate \ourmeth's scalability and sensitivity to various parameters. The experiment shows that 
our method's running time grows nearly linearly for both increasing number of instances and nodes. 
Additionally, we also investigate the sensitivity of \ourmeth to its three hyper-parameters, i.e. $\lambda_1$, $\lambda_2$ and $\lambda_3$. 
All combinations of hyperparamters exhibit similar robust behavior. Please find more details  at the supplement material.

We also investigate the sensitivity of \ourmeth to its three hyper-parameters:
$\lambda_1$, $\lambda_2$ and $\lambda_3$, which control the number of selected subnetwork, instances, and level of graphs smoothness in feature selection respectively.
These parameters all have a physical meaning and can provide useful control of the end user who want to enforce each of the corresponding behaviors based expert knowledge about the domain in which she employs \ourmeth.
However, in some cases the optimal parameter setting may be hard to obtain apriori, a typical challenge with many unsupervised methods. Hence we analyze the accuracy stability under small variations of these parameters. 

\begin{figure} [!t] 
    \centering
    \begin{subfigure}[t]{0.22\textwidth}
        \centering
        \includegraphics[width=\textwidth]{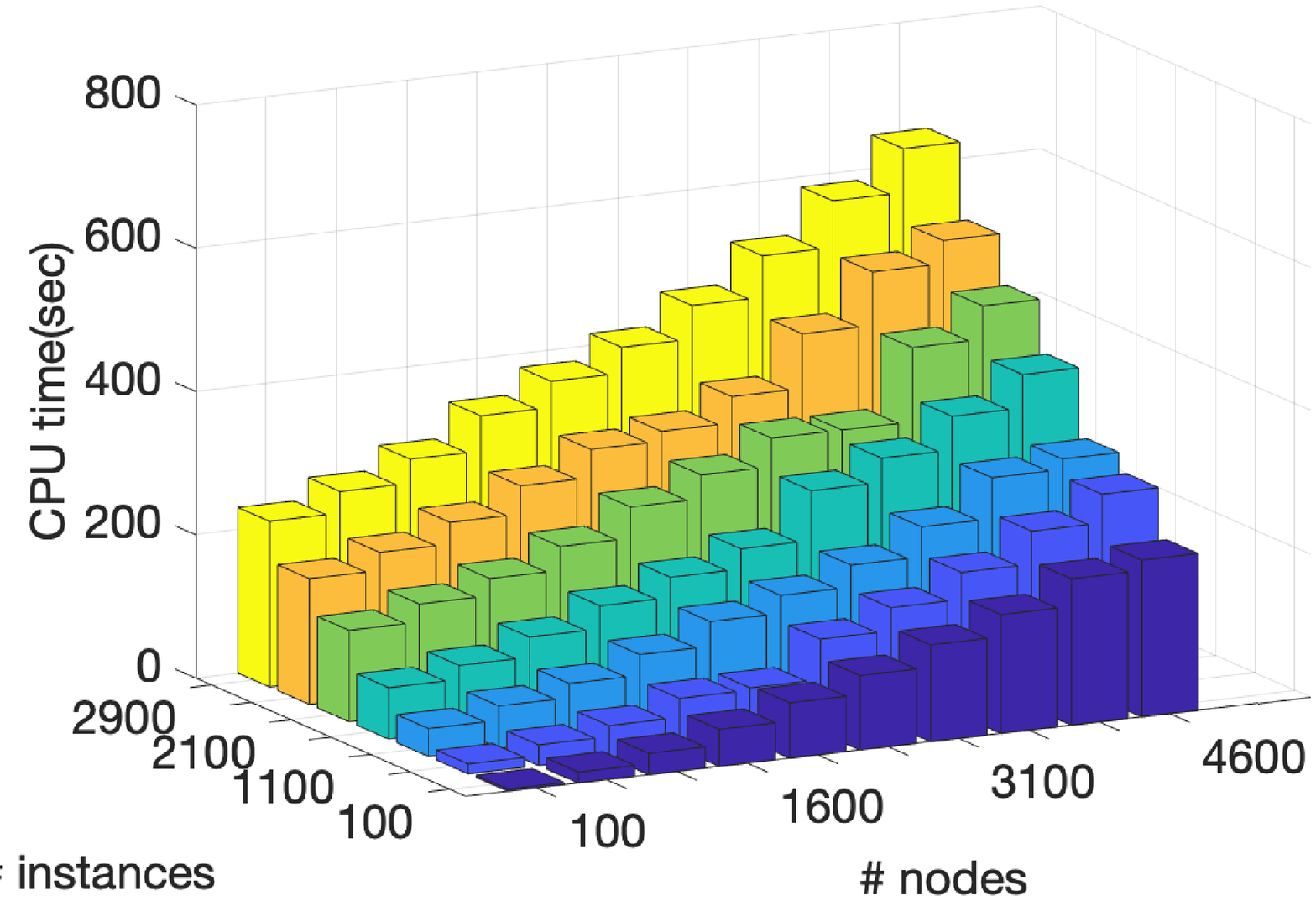}
        \caption{\footnotesize Running time}\label{fig:runing_time}
    \end{subfigure}
    \begin{subfigure}[t]{0.22\textwidth}
         \centering
         \includegraphics[width=\textwidth]{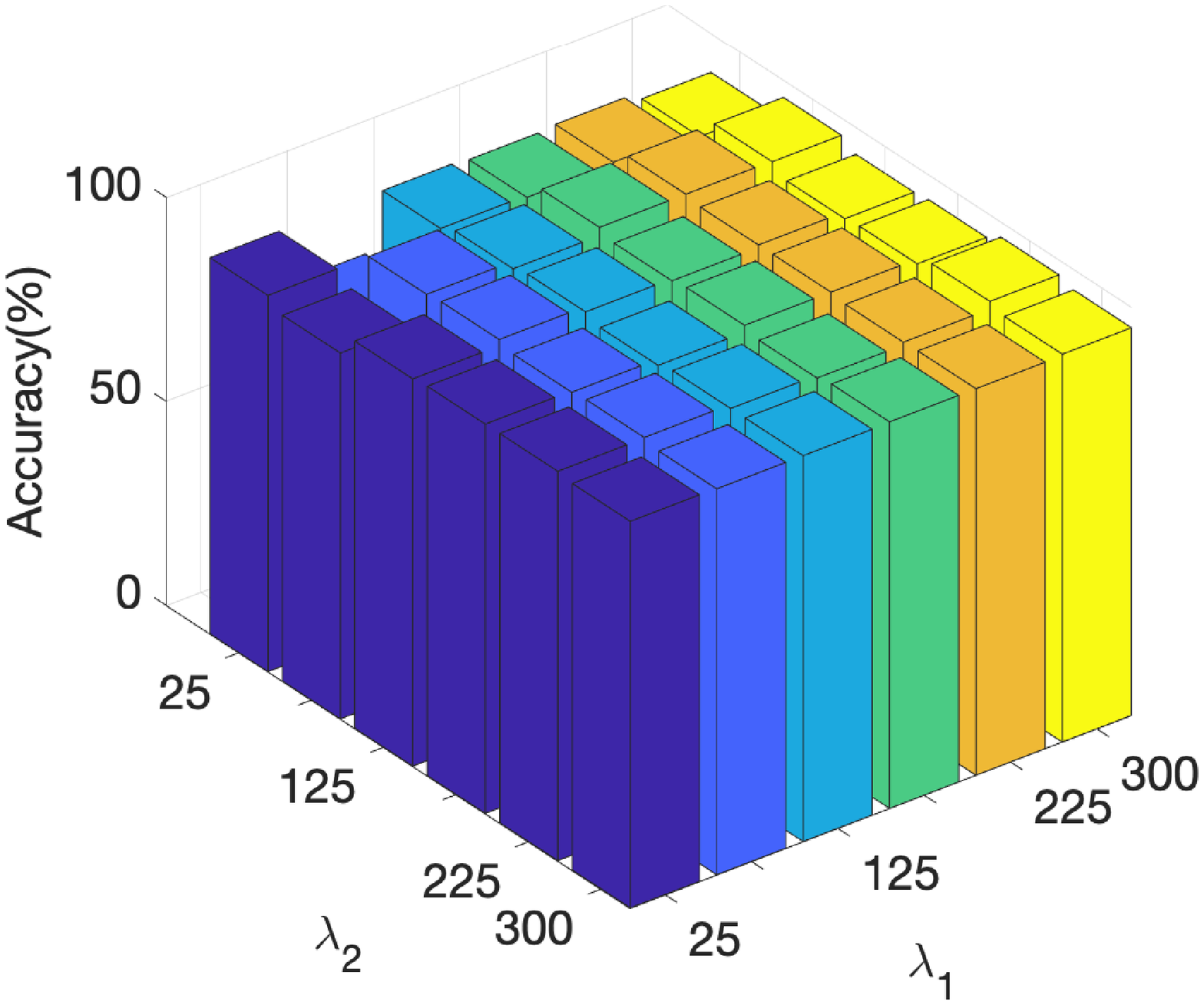}
         \caption{\footnotesize$\lambda_1$ v.s. $\lambda_2$}\label{fig:varying_lambda12}
     \end{subfigure}
    \begin{subfigure}[t]{0.22\textwidth}
        \centering
        \includegraphics[width=\textwidth]{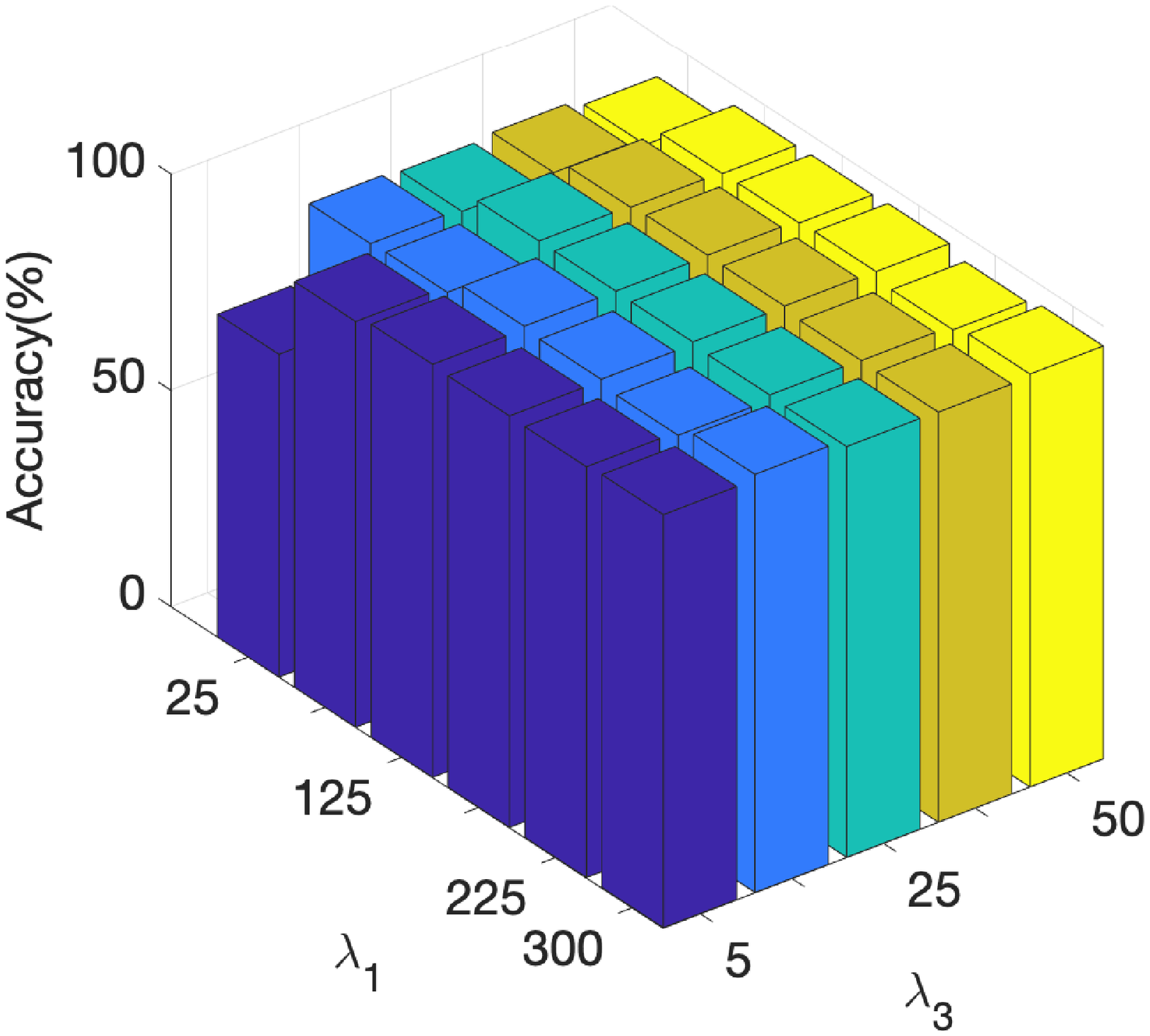}
        \caption{\footnotesize$\lambda_1$ v.s. $\lambda_3$}\label{fig:varying_lambda23}
    \end{subfigure}
    \begin{subfigure}[t]{0.22\textwidth}
        \centering
        \includegraphics[width=\textwidth]{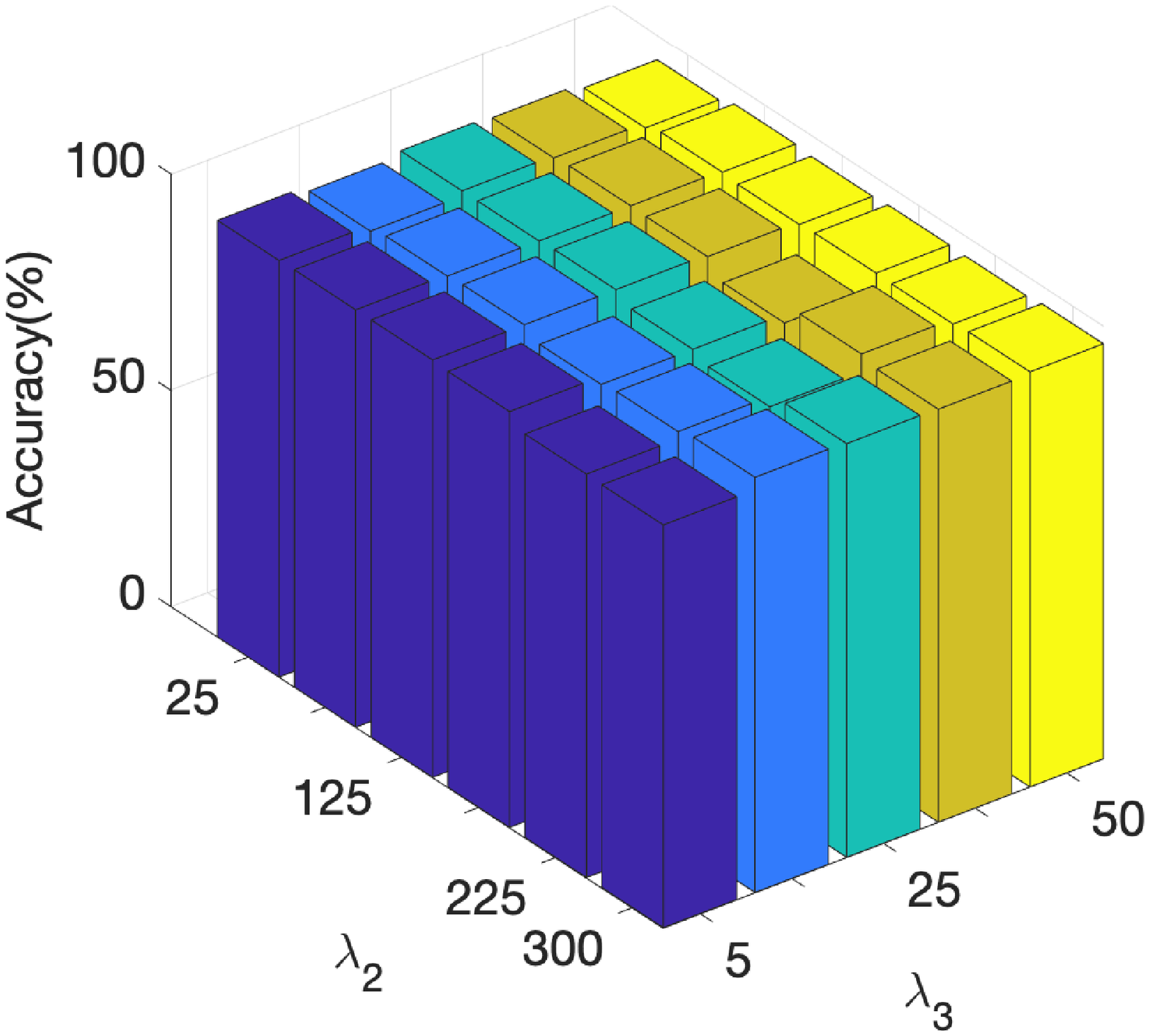}
        
        \caption{\footnotesize $\lambda_2$ v.s. $\lambda_3$}\label{fig:varying_lambda13}
    \end{subfigure}
    \caption{
    Parameter sensitivity and running time on the synthetic dataset.}
    \label{fig:param_sensitvity} \vsa
\end{figure}

Based on our analysis presented in Fig.~\ref{fig:param_sensitvity} the quality of \ourmeth does not vary significantly with the parameter settings.  
For this analysis, we measure the performance in terms of feature selection accuracy by fixing one parameter ($\lambda_3$) and updating the other two.
The overall performance on different combinations is stable and close to peak accuracy over large parameter ranges.   
The performance drops when at one of the parameter becomes significantly smaller (order of magnitude) than the other two. The reason for this behavior is that the three regularization terms in our objective have similar contribution to the cost, and thus keeping their importance balanced results in optimal performance.
In addition, small values of $\lambda_1$ and $\lambda_2$ will not enforce sufficient sparsity for the representative instance and feature selection and lead to poor performance due to negative effect of outliers and noise. 
Other combinations of hyperparamters exhibit similar robust behavior (in supplement material). Also, by default we set $k=10$, but our analysis does not render UISS sensitive to third internal dimension of the self-representative factorization.

\section{Conclusion}
In this paper we proposed and evaluated \ourmeth: a general unsupervised approach for joint subnetwork and instance selection in network data.
It performs interpretable subnetwork and instance selections via a self-representative factorization which enforces smoothness on the inter-feature network and robustness to outlier instances and noise.
\ourmeth dominated alternatives in both instance and subnetwork selection, often achieving its highest accuracy using significantly fewer features and instances than those employed by corresponding baselines.
Meanwhile, our method is able to discover interpretable subnetwork in the data such as city center localities of bike rental behavior distinguishing between workdays and weekends. 



\section*{Acknowledgment}
The work is supported in part by the NSF Smart and Connected Communities (SC\&C) CMMI grant 1831547 and in part by the National Institutes of Health through NIH grant R01DE027953.

{
\footnotesize
\bibliographystyle{abbrv}
\bibliography{references}
}
\end{document}